\documentclass[a4paper,fleqn]{cas-dc}

\usepackage[authoryear]{natbib}

\usepackage{inconsolata}
\usepackage{amsmath}
\usepackage{amssymb}
\usepackage{booktabs}
\usepackage{multirow}
\usepackage{bm}
\usepackage{algorithm}
\usepackage{algorithmicx}
\usepackage{algpseudocode}
\usepackage{subcaption}

\def\tsc#1{\csdef{#1}{\textsc{\lowercase{#1}}\xspace}}
\tsc{WGM}
\tsc{QE}

\begin{document}
\let\WriteBookmarks\relax
\def\floatpagepagefraction{1}
\def\textpagefraction{.001}
\let\printorcid\relax

\shorttitle{}    

\shortauthors{}  

\title [mode = title]{RuPLaR : Efficient Latent Compression of LLM Reasoning Chains with Rule-Based Priors From Multi-Step to One-Step}  

\author[suda]{Xiaocheng Luo}
\ead{xcluo@stu.suda.edu.cn}

\author[suda]{Kang Wang}
\ead{kwang0822@stu.suda.edu.cn}

\author[ece]{Zaifu Zhan}
\ead{zhan8023@umn.edu}

\author[suda]{Yuechi Zhou}
\ead{yczhouyc@stu.suda.edu.cn}

\author[suda]{Xiangyu Duan}
\ead{xiangyuduan@suda.edu.cn}
\cormark[1]
\cortext[1]{Corresponding author}

\affiliation[suda]{organization={School of Computer Science and Technology},%
    addressline={Soochow University}, 
    city={Suzhou},
    postcode={215006}, 
    state={Jiangsu},
    country={PR China}}

\affiliation[ece]{organization={Department of Electrical and Computer Engineering},%
    addressline={University of Minnesota}, 
    city={Minneapolis},
    postcode={55455}, 
    state={MN},
    country={USA}}


\begin{abstract}
    The Chain-of-Thought (CoT) paradigm, while enhancing the interpretability of Large Language Models (LLMs), is constrained by the inefficiencies and expressive limits of natural language. Latent Chain-of-Thought (latent CoT) reasoning, which operates in a continuous latent space, offers a promising alternative but faces challenges from structural complexities in existing multi-step or multi-model paradigms, such as error propagation and coordination overhead. In this paper, we  introduce \textbf{One-Model One-Step}, a novel compression framework for \textbf{La}tent \textbf{R}easoning with \textbf{Ru}le-Based \textbf{P}riors(\textbf{RuPLaR}) to address this challenge. Our method trains an LLM to autonomously generate latent reasoning tokens in a single training stage, guided by rule-based prior probability distributions, thereby eliminating cascaded processes and inter-model dependencies. To ensure reasoning quality, we design a joint training objective that enforces answer consistency via cross-entropy, aligns soft token distributions with rule-based priors via KL divergence (the Soft Thinking constraint), and adds a problem‑thought semantic alignment constraint in the representation space. Extensive experiments show that our compression framework not only improves accuracy by 11.03\% over existing latent CoT methods on GSM8k-Aug but also achieves this with minimal token usage, underscoring its effectiveness and extensibility. Code: \href{https://github.com/xiaocen-luo/RuPLaR}{https://github.com/xiaocen-luo/RuPLaR}.
\end{abstract}





\maketitle

\section{Introduction}

The Chain-of-Thought (CoT) paradigm \citep{wei2022chain}, which elicits Large Language Models (LLMs) to articulate step-by-step reasoning in natural language, has significantly enhanced the interpretability and performance of LLMs on complex tasks \citep{guo2025deepseek,chen2025towards,chen2025reasoning}. By externalizing intermediate thought processes, CoT provides a transparent window into the model's decision-making and enables human oversight. 

However, the very medium of natural language imposes fundamental constraints on the reasoning capacity of LLMs. The primary limitation is concept–symbol inefficiency: tokens that carry essential reasoning concepts (e.g., ``therefore'', ``necessarily'') are locked into grammatically required but functionally redundant structures (e.g., ``of'', ``the''), increasing token usage and inference cost without adding reasoning value \citep{feng2025efficient,sui2025stop}. A second constraint is the semantic bottleneck: human cognition often involves abstract, continuous, or multi-concept representations that transcend discrete linguistic symbols and are difficult to verbalize precisely \citep{wittgenstein1921tractatus}. For instance, solving ``30\% of 600'' is immediate for humans as a continuous magnitude estimation, but LLMs must generate discrete token sequences like \texttt{600*30/100=180} step by step.

These inherent limitations have motivated a shift from explicit language chains towards \textbf{Latent Chain-of-Thought (latent CoT)} reasoning, where reasoning operates within a continuous latent space \citep{zhang2025latentevolve}. Latent CoT models replace discrete tokens with high-density, high-dimensional latent representations, which serve as a more natural and efficient substrate for computational reasoning. Nonetheless, this emerging paradigm introduces new core challenges stemming from the unobservable nature of the reasoning process. This opacity creates significant training and coordination difficulties, making direct supervision hard to implement \citep{xu-etal-2025-softcot,ruan2025reasoning}.

\begin{figure}
	\centering
	\includegraphics[width=0.96\linewidth]{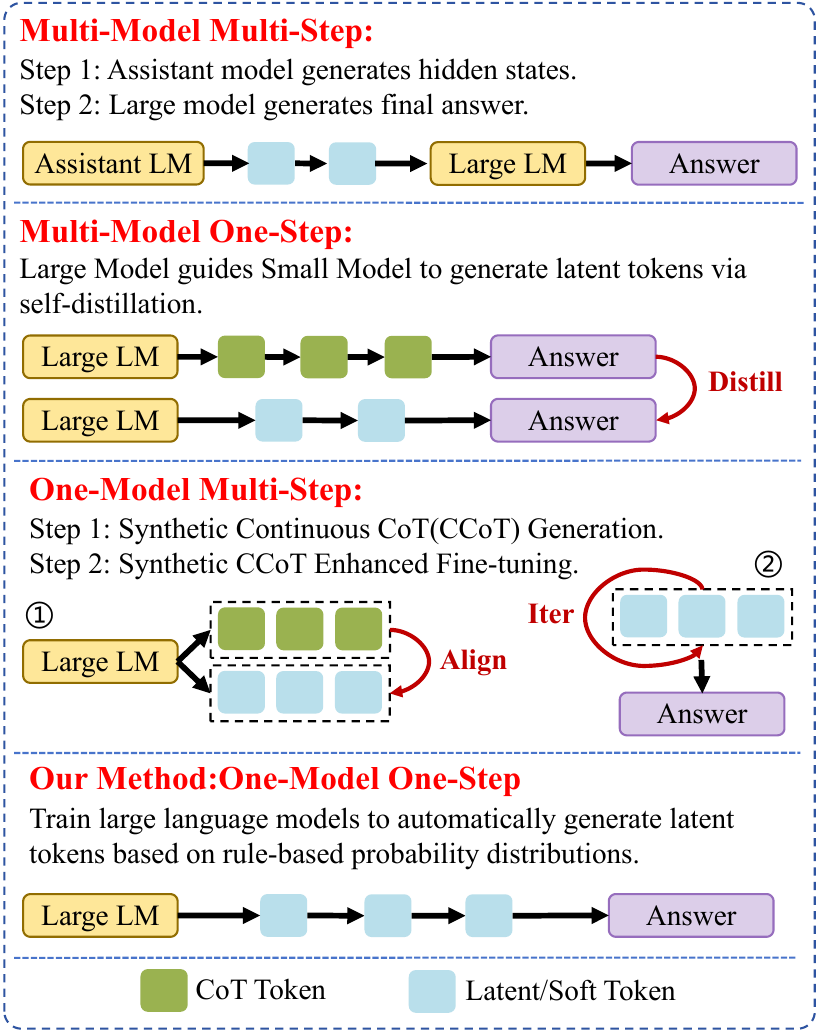}
	\caption{\label{fig:latent reasoning} The latent reasoning paradigms of existing approaches and our method. Note that multiple latent reasoning paradigms exist in existing methods; only the most typical paradigm is selected as representative.}
\end{figure}

To address these challenges, research on latent CoT reasoning has undergone rapid and vigorous development. As illustrated in Figure~\ref{fig:latent reasoning}, existing methodologies can be broadly classified into three paradigms according to their architectural and procedural designs: \emph{Multi-Model Multi-Step}, \emph{Multi-Model One-Step}, and \emph{One-Model Multi-Step}. The first paradigm \citep{xu-etal-2025-softcot,wang2025lta,he2025semcot,xu2025softcot++} employs a cascaded framework in which an assistant model generates soft reasoning traces, which are subsequently utilized by an LLM to produce the final answer. The second paradigm \citep{shen2025codi,kuzina2026kava} utilizes distillation techniques to condense discrete explicit CoT reasoning into dense latent tokens in a single forward pass of the assistant model, making it one-step at the paradigm level. The third paradigm \citep{wu2025parallel,tan2025think,liu2025marcos} encompasses diverse methodological strategies. Among these, one representative approach \citep{wang2025synadapt} involves aligning latent tokens with explicit reasoning chains, followed by iterative refinement of these latent tokens.

While these paradigms have demonstrated promise, they are inherently limited by their structural complexities. Multi-step approaches, both within and across models, often suffer from error propagation, where inaccuracies in early stages compromise the final outcome. Meanwhile, multi-model frameworks face challenges in achieving tight collaboration between components, leading to suboptimal information flow and increased computational overhead. It is worth clarifying that in this taxonomy, ``step'' refers to a distinct training stage. A ``one-step'' method, consequently, means the model is trained end‑to‑end in a single stage, without cascaded pre‑training or separate auxiliary model training. Such a method may still generate multiple dense tokens autoregressively during inference—this is not a ``multi-step'' process in the training-stage sense, but rather the standard autoregressive generation procedure for sequence models.

Inspired by the above insights, we introduce \emph{One-Model One-Step}, a novel compression paradigm that embodies this principle. Here, ``one‑model'' means a single LLM handles both latent reasoning and answer generation without auxiliary models. ``One‑step'' (as just defined) means single‑stage training with a joint loss, eliminating multi‑stage pipelines. By design, this unified architecture inherently circumvents the error propagation issue that plagues multi-step approaches; it also obviates the need for inter-model coordination, thereby avoiding the suboptimal information flow and additional computational overhead associated with multi-model frameworks. Concretely, our method trains an LLM to autonomously generate latent reasoning tokens—autoregressively producing a small number (typically 2–3) of such tokens per instance—guided by rule‑based probability distributions. This integrated approach eliminates cascaded steps and inter-model coordination, mitigating error propagation and fostering a more cohesive reasoning process within a unified model.

This paper addresses a specific, under-explored question: Can we construct a more stable and interpretable supervision signal for latent reasoning using explicit structural information from the reasoning chain itself? Our key insight is that in structured reasoning tasks (e.g., arithmetic, symbolic temporal reasoning), each reasoning step has a well-defined input-output structure. By extracting operational tokens (inputs/operators) and result tokens (outputs) from each explicit reasoning step, we can construct rule-based prior distributions that directly encode the semantics of the step. These priors provide a cleaner, more stable training signal than learned alternatives, eliminating the need for auxiliary models or reinforcement learning while maintaining differentiability for gradient-based optimization.

To ensure the quality of latent CoT reasoning, we employ a joint training objective comprising three key components. The first is the \textbf{Answer Consistency Constraint}: implemented as a standard cross-entropy loss, this ensures the accuracy and coherence of the final output. The second is the \textbf{Soft Thinking Paradigm Constraint} (see Section~\ref{Related Works}): formulated as a KL divergence loss, this term minimizes the divergence between the rule-based prior probability distribution and the distribution represented by each generated soft token distribution. The third is the \textbf{Problem-Thought Semantic Constraint}: also implemented as a KL divergence loss, this aligns the semantic distribution of generated thinking tokens with that of the input question token, ensuring the latent reasoning path remains grounded in the original problem context.
To demonstrate its effectiveness, we evaluate our model on four mathematical reasoning benchmarks: GSM8k-Aug, GSM-Hard, SVAMP, and MultiArith. Extensive experiments show that it achieves state-of-the-art performance with minimal reasoning length, confirming the strength of latent reasoning.

In summary, our main contribution is four-fold:

\begin{itemize}
    \item We propose training LLMs using rule-based prior probability distributions as supervision. To our knowledge, this direction has received limited exploration in the latent reasoning literature, and offers a promising alternative to training-derived priors that may introduce noise.
    \item We propose a novel \emph{One-Model One-Step} training framework as a variant of the trainable Soft Thinking paradigm. This framework reduces system complexity, computational overhead, and inference latency compared to \emph{Multi-Step} or \emph{Multi-Model} alternatives.
    \item We devise a focused KL-divergence loss that aligns the model's generated soft token distributions with Rule-Based Priors, applied only to semantically critical tokens. This ensures the core reasoning content is preserved without over-constraining the rest.
    \item Extensive experiments validate the superiority of our method, which not only achieves a 11.03\% accuracy gain over existing latent reasoning approaches but also attains this result with the minimum token consumption.
\end{itemize}

\section{Related Works}
\label{Related Works}

Our work focuses on latent reasoning. We categorize it into three primary groups based on the origin of the latent representation: the model's internal \underline{hidden states}, \underline{weighted embeddings} of existing vocabulary tokens, and newly introduced \underline{special vectors}.

\textbf{Hidden State.} COCONUT \citep{hao2025training} pioneered using the final hidden state as input for latent reasoning, though it suffers from spatial inconsistency between hidden states and token embeddings. CODI \citep{shen2025codi} addresses this by incorporating an MLP for mapping and knowledge distillation. PCCOT \citep{wu2025parallel} further improves efficiency via Jacobi iteration, while KaVa \citep{kuzina2026kava} leverages KV cache distillation. An alternative approach employs auxiliary models to generate latent thoughts \citep{xu-etal-2025-softcot,zhang2025latentevolve,wang2025lta}. In contrast, our method directly operates on the input token embedding space, avoiding the spatial mismatch between hidden states and token embeddings.

\textbf{Weighted Embedding.} Soft Thinking \citep{zhang2025soft} and MoI \citep{zhuang2025mixture} combine token probability distributions with embeddings in a training-free manner. Subsequent work enables training through reinforcement learning \citep{butt2026soft,zheng2025soft,yue2025hybrid}. CoLaR \citep{tan2025think} introduces latent heads to predict compressed embeddings. Our method extends the Soft Thinking paradigm into a trainable framework that significantly reduces token usage while enabling multi-path exploration.

\textbf{Special Vector.} DART \citep{jiang2025dart} and SynAdapt \citep{wang2025synadapt} employ dedicated vector parameters through knowledge distillation and synthetic CoT generation. Other approaches insert latent tokens directly into input \citep{sun2025enhancing} or use complex learned states \citep{liu2025marcos,zheng2025fast}. For representation learning, Token Assorted \citep{su2025token} uses VQ-VAE for quantization, while CoCoMix \citep{tack2026llm} employs sparse autoencoders. ReGuLaR \citep{Wang2026ReGuLaR} formulates latent reasoning within a Variational Auto-Encoding (VAE) framework.

\section{Problem Definition and Notations}

This section formalizes the reasoning task and introduces our notation. We focus on mathematical reasoning as a representative testbed for our compression framework. The framework is designed for reasoning domains where explicit, segmentable symbolic structures are available—such as arithmetic expressions, symbolic temporal operations, or program traces—which enable the construction of rule-based priors via step segmentation and token extraction. For less structured domains such as commonsense or narrative reasoning, where explicit symbolic intermediates are absent, our current prior construction pipeline is not directly applicable (see Limitations). While the core idea of distilling reasoning into compact latent representations is conceptually general, the current work focuses on mathematical reasoning as the primary testbed, with a preliminary experiment on symbolic temporal reasoning (Date Understanding). Comprehensive validation on other structured tasks, and extensions to less structured domains, remain future work.

The task is defined on a dataset \(\mathcal{D}\), where each instance comprises a question \(\mathbf{t}_q = (t_{q}^{1}, \dots, t_{q}^{L_q})\), a reasoning chain \(\mathbf{t}_r = (t_{r}^{1}, \dots, t_{r}^{L_r})\), and an answer \(\mathbf{t}_a = (t_{a}^{1}, \dots, t_{a}^{L_a})\), with \(L_q, L_r, L_a\) denoting their respective token lengths after standard tokenization. A representative example entry would be: \emph{``Question: Out of 600 employees in a company, 30\% got promoted while 10\% received bonus. How many employees did not get either a promotion or a bonus?''}, \emph{``Reasoning chain: \texttt{<<600 * 30 / 100 = 180>> <<600 * 10 / 100 = 60>> <<180 + 60 = 240>> <<600 - 240 = 360>>}''},\footnote{Each arithmetic expression enclosed in \texttt{<<} and \texttt{>>}, e.g., \texttt{<<600 * 30 / 100 = 180>>}, is tokenized into multiple discrete tokens (digits, operators, and the equals sign) following the model's standard tokenizer. The delimiters \texttt{<<} and \texttt{>>} are used to segment reasoning steps and are not part of the actual arithmetic expression.} and \emph{``Answer: 360''}.

In conventional multi-step reasoning, the intermediate chain \(t_r\) is often long and token-inefficient, slowing down inference and increasing memory overhead. To overcome this, we propose \textbf{reasoning compression}—mapping lengthy reasoning sequences into a compact latent space while preserving semantic fidelity. At the core of this approach is the notion of soft tokens, defined as follows:

\textbf{Soft Token Representation.} \label{def:soft_token}
We introduce a hierarchical terminology to distinguish two related but distinct concepts:

\begin{itemize}
    \item \textbf{Soft Token Distribution} (\(\pi_i \in \Delta^{|V|}\)): A probability distribution over the vocabulary \(V\), where \(\Delta^{|V|}\) denotes the probability simplex. This distribution encodes the model's ``soft'' choice of tokens for a reasoning step. Its role differs across phases:
    \begin{itemize}
        \item During \textbf{data preprocessing}, for each reasoning step \(r_i\) extracted from the explicit chain-of-thought, we construct a corresponding target distribution \(\pi_i\) using rule-based methods (see Figure~\ref{fig:our model}, left panel). These \(\pi_i\)'s serve as training targets.
        \item During \textbf{training and inference}, \(\pi_i\) denotes the distribution \textit{generated by the model} for the \(i\)-th reasoning step, which is trained to align with the target distribution (during training) or used to derive a soft token embedding (during inference).
    \end{itemize}

    \item \textbf{Soft Token Embedding} (or \textbf{Soft Token}) (\(\mathbf{z}_i \in \mathbb{R}^d\)): A continuous vector derived from a soft token distribution \(\pi_i\). It is computed as a convex combination of vocabulary embeddings:
    \begin{equation}
        \mathbf{z}_i = \sum_{j=1}^{|V|} \pi_{i,j} \, \mathbf{e}_j, \label{equ:soft_token}
    \end{equation}
    where \(\pi_{i,j}\) is the \(j\)-th component of the distribution \(\pi_i\), and \(\mathbf{e}_j \in \mathbb{R}^d\) is the embedding of the \(j\)-th vocabulary token.
\end{itemize}

Throughout the paper, we use \textit{soft tokens} (plural) to refer to the sequence \(\mathbf{z}_{1:N}\) generated during reasoning, where each individual \(\mathbf{z}_i\) is a \textit{soft token} and \(N\) denotes the number of reasoning steps in the chain-of-thought. The term \textit{soft token distribution} is used when referring to the probability distribution \(\pi_i\) rather than the resulting embedding.

\section{Method}
\label{sec:method}

\begin{figure*}[!ht] 
	\centering
	\includegraphics[width=0.96\linewidth]{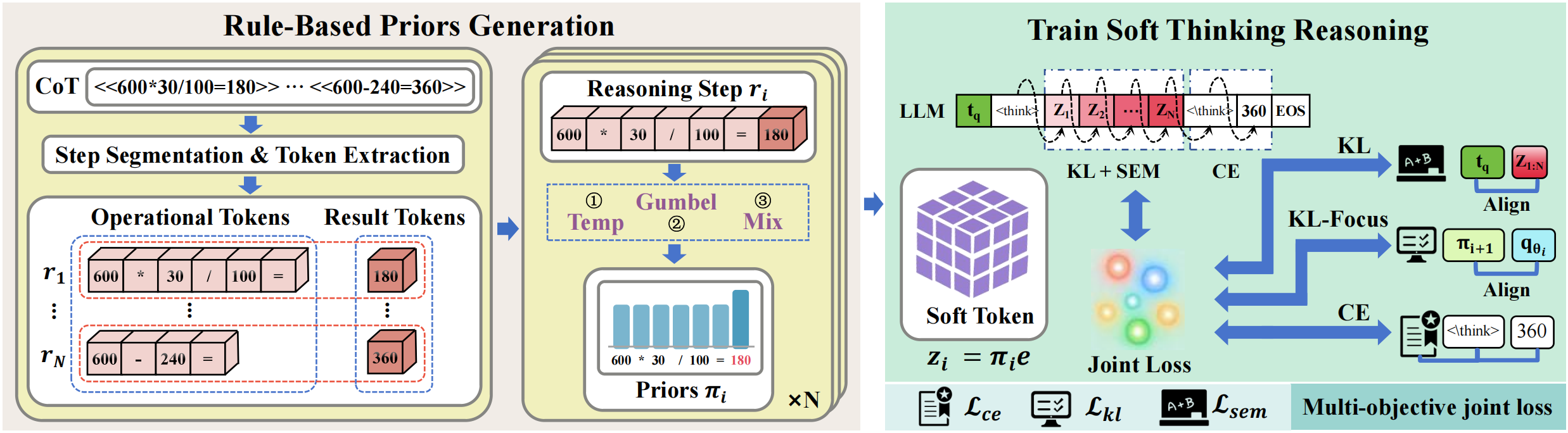}
    \caption{\label{fig:our model} Overview of our model. Our model operates in two distinct phases: (a) during data preprocessing, rule-based methods generate the target soft token distributions \(\pi_i\) for each reasoning step \(r_i\); and (b) during training, the LLM learns to produce soft token distributions that align with these targets, from which soft token embeddings \(\mathbf{z}_i\) are derived. The Rule-based Coefficient Generation component (left) presents three methodological variants (indicated by dashed borders) for constructing target distributions. The red rounded rectangle represents the output of Step Segmentation, while the blue rounded rectangle represents the output of Token Extraction.}
\end{figure*}

To address the challenge of lengthy reasoning sequences, we introduce a compression-based framework with \emph{One-Model One-Step} that maps multi-step reasoning into a compact latent space, thereby enhancing the efficiency of LLMs. This approach is designed to fulfill three primary objectives:
(1) to encode reasoning tokens into dense latent representations and enable the model to interpret these compressed states;
(2) to autoregressively predict subsequent latent states and decide when to halt reasoning; and
(3) to facilitate latent reasoning exploration and exploitation.

Achieving these goals necessitates first constructing target soft token distributions (Section~\ref{def:soft_token}) as supervision signals for model training. This is accomplished in a two-stage process: First, during data preprocessing, rule-based methods construct prior distributions \(\pi_i\) for each reasoning step \(r_i\) as training targets. Second, the model is trained to align its generated distributions with these targets, from which soft token embeddings \(\mathbf{z}_i\) are derived. The overall framework is illustrated in Figure~\ref{fig:our model}.

\subsection{Phase 1: Data Preprocessing — Rule-Based Priors Generation}

To train the model for latent reasoning, we first need to establish target soft token distributions (\(\pi_i\)) that encode compressed semantic information for each reasoning step. As direct supervision from human-annotated soft token distributions is unavailable, we generate Rule-Based Priors through a three-stage process: (1) segmentation of the explicit reasoning chain into computational steps, (2) token extraction from each step, and (3) differentiable prior construction. This approach transforms discrete reasoning chains into continuous probability distributions while preserving essential arithmetic semantics.

To transform the explicit reasoning chain \(\mathbf{t}_r\) into a differentiable learning signal, we decompose it into individual computational steps. Formally, for each reasoning step \(r_i\) (e.g., \texttt{20 * 1.20 = 24}), we construct a prior distribution \(\pi_i \in \mathbb{R}^{|V|}\) over the vocabulary \(V\), where \(\pi_i\) is a probability vector (i.e., \(\pi_{i,j} \geq 0\) and \(\sum_{j=1}^{|V|} \pi_{i,j} = 1\)). This prior defines a convex combination of vocabulary embeddings, consistent with the soft token distribution definition in Section~\ref{def:soft_token}. The distribution is designed to concentrate probability mass on semantically relevant tokens while maintaining differentiability for gradient-based learning.

\subsubsection{Step Segmentation and Token Extraction}

We decompose each explicit reasoning chain into discrete steps based on domain-specific structural cues. For mathematical reasoning, we use delimiters (e.g., \texttt{<<...>>}) that naturally segment computational substeps. Unlike fixed-length segmentation approaches \citep{tan2025think}, our method adapts dynamically to the complexity of each reasoning chain, allowing a variable number of steps per problem. For each extracted step $r_i$, we identify two generic token categories that are instantiated differently depending on the domain:

\begin{itemize}
    \item \textbf{Operational Tokens:} Tokens that represent the input elements of the step (e.g., in mathematical reasoning: numerical values, operators, and constants). For instance, in \texttt{20 * 1.20 = 24}, these include \texttt{20}, \texttt{*}, \texttt{1.20}.
    \item \textbf{Result Tokens:} The token(s) that represent the output of the step (e.g., \texttt{24} in the same example).
\end{itemize}

These categories serve as domain-specific instantiations of a more general principle: any structured reasoning step can be characterized by its input elements and its output. Extending our framework to other tasks with similar explicit symbolic structures (e.g., program execution or symbolic temporal reasoning) would require adapting the segmentation cues and token categories to the conventions of those domains. However, for tasks where reasoning is less structured—such as commonsense reasoning or open-ended narrative reasoning—the lack of explicit symbolic intermediates poses a more fundamental challenge (see Limitations). 

This adaptive tokenization scheme preserves the essential arithmetic logic of multi-step reasoning while accommodating natural variations in problem complexity. By supporting dynamic segmentation, our approach provides a flexible yet structured foundation for subsequent soft token representation learning.

Before proceeding to prior construction, we address a practical tokenization consideration that affects all downstream operations.

\textbf{Tokenization-Aware Prior Construction.}
A practical consideration is that numerical values and arithmetic expressions may be split into multiple tokens by the tokenizer (e.g., the value ``24'' could be tokenized as a single token or as separate tokens ``2'' and ``4'', depending on the vocabulary). Our implementation handles this transparently: when extracting operational tokens and result tokens, we first extract the raw substrings (e.g., ``24'', ``*'', ``1.20''), then encode each substring through the model's tokenizer. All token IDs produced from a given substring are assigned the same logit value ($\beta_{\text{op}}$ for operational substrings, $\beta_{\text{res}}$ for the result substring). Consequently, the prior distribution naturally distributes probability mass across all sub-token representations of a numerical value, preserving semantic fidelity without requiring manual token grouping. Repeated tokens are handled idempotently; assigning the same logit multiple times has no additional effect.

\subsubsection{Prior Construction Methods}

Following token extraction, we transform the discrete tokens into continuous probability distributions suitable for gradient-based learning. The core challenge lies in balancing semantic specificity with training stability—ensuring the prior concentrates on reasoning-relevant tokens while remaining numerically robust for optimization. We address this through three complementary approaches, each offering distinct trade-offs between focus, smoothness, and differentiability:

(1) \textbf{Temperature-based Prior Construction.} We construct the prior distribution \(\pi_i\) for each reasoning step \(r_i\) through a temperature-scaled categorical distribution defined over the vocabulary \(V\). Formally,
\begin{equation}
    p_{\text{Temp}} = \pi_i = p(\cdot \mid r_i) = \mathrm{Softmax}\left(\frac{\bm{\ell}}{\tau}\right), \label{equ:Temp}
\end{equation}
where \(\bm{\ell} \in \mathbb{R}^{|V|}\) is a logit vector. To encode the relevance of extracted tokens, we initialize \(\ell_j = -\infty\) for all tokens \(j\) in the vocabulary \(V\) not appearing in the reasoning step. For tokens in the operational set \(\mathcal{K} = \{j_1, \dots, j_m\}\), we set \(\ell_j = \beta_{\text{op}}\), while for the result token \(j_{\text{res}}\), we assign a higher value \(\ell_{j_{\text{res}}} = \beta_{\text{res}}\) with \(\beta_{\text{res}} > \beta_{\text{op}}\). The temperature hyperparameter \(\tau > 0\) governs the entropy of the distribution: as \(\tau \to 0\), the distribution approaches a deterministic focus on the highest-logit tokens.

(2) \textbf{Gumbel-Softmax Prior for Differentiable Sampling.} For scenarios requiring stochastic latent variables, we employ Gumbel-Softmax relaxation to enable differentiable categorical sampling for each reasoning step \(r_i\):
\begin{equation}
    p_{\text{Gumbel}} = \mathrm{GumbelSoftmax}(\bm{\ell}, \tau, \mathrm{hard}=False),
\end{equation}
where the operation is defined as:
\begin{equation}
    p_j = \frac{\exp((\ell_j + g_j)/\tau)}{\sum_{k=1}^{|V|} \exp((\ell_k + g_k)/\tau)}, \quad 
    g_j \sim \mathrm{Gumbel}(0,1),
\end{equation}
where \(j\) indexes over the vocabulary \(V\). The parameter \(\mathrm{hard}\) selects between soft relaxation (training) and discrete approximation (inference), maintaining gradient flow while supporting stochastic reasoning representations.

(3) \textbf{Mixture Prior for Balanced Token Emphasis.} To explicitly balance attention between computational process and final outcome, we design a mixture prior:
\begin{equation}
    p_{\text{mix}} = \lambda \cdot p_{\text{uniform}} + (1-\lambda) \cdot p_{\text{result}},
\end{equation}
where \(p_{\text{uniform}}\) distributes probability uniformly over operational tokens \(\mathcal{K}\):
\begin{equation}
    p_{\text{uniform}}(j) = \begin{cases}
        \frac{1}{|\mathcal{K}|}, & \text{if } j \in \mathcal{K} \\
        0, & \text{otherwise}
    \end{cases},
\end{equation}
and \(p_{\text{result}}\) concentrates on the result token \(j_{\text{res}}\):
\begin{equation}
    p_{\text{result}}(j) = \mathbb{I}[j = j_{\text{res}}].
\end{equation}
The mixing coefficient \(\lambda \in [0,1]\) provides explicit control over the trade-off between modeling the computational procedure (\(\lambda \to 1\)) and emphasizing the final answer (\(\lambda \to 0\)).

These three methods offer different trade-offs between semantic focus, training stability, and computational properties, allowing flexible adaptation to varying training requirements while maintaining the core objective of encoding reasoning semantics into differentiable priors. More details about Rule-Based Priors Construction Pipeline in Appendix \ref{app:rule_based_priors}.

\subsubsection{Why Rule-Based Priors? A Comparative Perspective}
\label{sec:why_rule_based}

\begin{table*}[!ht]
    \centering
    \caption{Comparison of supervision strategies for latent Chain-of-Thought reasoning.}
    \label{tab:prior_comparison}
    \begin{tabular}{p{2.8cm}p{3.2cm}p{3.8cm}p{4.2cm}}
    \toprule[2pt]
    \textbf{Supervision Type} & \textbf{Examples} & \textbf{Limitations} & \textbf{Our Advantage} \\
    \midrule
    Learned priors & A data-driven prior based on token position statistics & Noisy, unstable, requires extensive tuning; prior must be inferred indirectly from downstream rewards or reconstruction loss & Directly grounded in explicit reasoning structure; stable and interpretable \\
    \hline
    Random priors & Randomly assigned token importance & No semantic signal; model cannot learn meaningful latent representations & Provides structured inductive bias through token extraction \\
    \hline
    Uniform priors & Assigning equal importance to all tokens & Treats all tokens equally; dilutes structural information in reasoning chains & Differentiated importance via operational/result token distinction \\
    \hline
    Result-Only Prior & Supervision using only the result token of each reasoning step & Ignores procedural semantics (operators, constants); loses information about the computation process & Captures both procedural (operational tokens) and outcome (result token) information \\
    \hline
    Operational-Only Prior & Supervision using only operational tokens, excluding the result token & Lacks outcome information; model cannot learn to predict the correct final result & Balances process and outcome via mixture coefficient \\
    \bottomrule[2pt]
    \end{tabular}
\end{table*}

To justify our design choice, we compare rule-based priors against five alternative supervision strategies for latent CoT, summarized in Table~\ref{tab:prior_comparison}. Our rule-based approach offers three concrete benefits: stability, interpretability, and computational efficiency. First, because the prior is constructed deterministically from the reasoning chain, training does not suffer from the convergence issues or reward shaping complexities common in learned-prior approaches. Second, the prior distribution directly reflects the content of each reasoning step—operational tokens encode how the computation proceeds, while the result token encodes what the step produces—and this transparency facilitates debugging and analysis (Section~\ref{sec:analysis}). Third, the focused KL-divergence reduces alignment cost from O(|V|) to O(k), whereas learned-prior methods often require full-distribution modeling or auxiliary networks.

Empirically, the advantages of our design are validated by the ablation results in Section~\ref{sec:ablation}. Using only the result token (Result-Only Prior) achieves 38.1\% average accuracy—a 13.3-point drop from the full model—while using only operational tokens (Operational-Only Prior) yields the same 38.1\%, indicating that neither signal alone is sufficient. In contrast, the full model, which combines both via the mixture coefficient \(\lambda\), substantially outperforms all single-category priors, confirming that procedural semantics and outcome signals provide complementary information essential for robust reasoning. Random, uniform, and learned priors similarly fall short, underscoring the necessity of our structured rule-based prior.

We emphasize that these advantages are most pronounced in domains with explicit symbolic structure. For less structured domains, the construction of meaningful priors remains an open challenge (see Limitations).

\subsection{Phase 2: Training — Joint Optimization of Latent Reasoning}

The training of our soft-thinking reasoning model employs a joint optimization framework that simultaneously ensures answer accuracy, prior distribution alignment, and semantic consistency. As illustrated in Figure \ref{fig:our model} (right), our approach directly optimizes the latent reasoning process through three complementary loss components, each addressing a specific aspect of the reasoning task.

\subsubsection{Answer Consistency Constraint}

To ensure the model produces correct final answers, we employ a standard cross-entropy loss on the output sequence:

\begin{equation}
    \mathcal{L}_{\text{ce}} = -\frac{1}{L_a}\sum_{i=1}^{L_a} \log p(t_a \mid t_q, \mathbf{z}_{1:N}),
\end{equation}
where \(t_a\) is the ground truth answer, \(t_q\) is the input question, and \(\mathbf{z}_{1:N}\) represents the sequence of generated soft tokens. This component guarantees that the compressed reasoning process ultimately leads to accurate solutions.

\subsubsection{Soft Thinking Paradigm Constraint}

We introduce a focused KL-divergence loss that aligns the model's generated soft token distributions with Rule-Based Priors on semantically critical tokens. Unlike standard approaches that enforce alignment across the entire vocabulary, our method concentrates computation on the most relevant tokens, reducing complexity from \(O(|V|)\) to \(O(k)\) while filtering gradient noise and preserving semantic specificity. This selective mechanism maintains fidelity to reasoning-essential vocabulary items, offering a principled alternative to exhaustive alignment strategies.

Given a generated soft token distribution \(q_{\theta}(\pi_i \mid \cdot)\) from the LLM for the \(i\)-th reasoning step, and its corresponding rule-based prior \(p_{\text{prior}}(\pi_i)\), we compute:
\begin{equation}
    \mathcal{L}_{\text{kl}} = \frac{1}{N}\sum_{i=1}^{N} \sum_{j \in \mathcal{T}_i} p_{\text{prior}}(j) \left[ \log p_{\text{prior}}(j) - \log q_{\theta}(j) \right], \label{equ:kl}
\end{equation}
where \(\mathcal{T}_i\) denotes the set of vocabulary tokens in \(p_{\text{prior}}(\pi_i)\) with probability above threshold \(\delta\). The teacher distribution \(p_{\text{prior}}\) is normalized over the full vocabulary prior to thresholding; the mask \(\mathcal{T}_i\) filters out low-probability tokens during loss computation. The gradient only flows through the selected tokens, reducing the effective computational cost of the KL term from O(|V|) to O(k) per alignment step, while the forward pass remains unchanged. This selective focus preserves semantic fidelity while mitigating gradient noise from irrelevant vocabulary tokens.

\subsubsection{Problem-Thought Semantic Constraint}

While the prior alignment constraint ensures mathematical correctness at the distribution level, we further introduce a semantic coherence constraint that operates in the representation space to enforce logical flow and information consistency throughout the reasoning chain. This constraint addresses a fundamental aspect of coherent reasoning: problem-thought semantic alignment, which ensures that each reasoning step remains relevant to the original question.

Formally, let \(\mathbf{h}_q \in \mathbb{R}^{d}\) denote the hidden representation of the question \(\mathbf{t}_q\) at its final token position, obtained from the LLM's encoder. For each generated soft token \(\mathbf{z}_i\) (defined in Section~\ref{def:soft_token}) in the latent reasoning path of length \(N\), we obtain its corresponding hidden representation \(\mathbf{h}_{z_i} \in \mathbb{R}^{d}\) for \(i = 1, \dots, N\). To enforce semantic alignment between each soft token and the original question, we treat each of the \(d\) hidden dimensions as a category and minimize the Kullback–Leibler divergence between their softmax-normalized distributions:

\begin{equation}
    \mathcal{L}_{\text{sem}} = \frac{1}{N} \sum_{i=1}^{N} D_{\text{KL}}\left(\text{Softmax}(\mathbf{h}_q) \parallel \text{Softmax}(\mathbf{h}_{z_i})\right)
\end{equation}

This formulation provides a principled, scale-invariant measure of distributional similarity in the representation space. Unlike MSE, which is sensitive to feature scale and can be difficult to tune across layers, or cosine similarity, which lacks a probabilistic grounding, the KL divergence over softmax-normalized feature dimensions naturally enforces that the latent thought representations maintain a similar distributional profile to the question representation, anchoring each reasoning step in the original problem context. We also considered cross-attention as an alternative alignment mechanism, but it introduces additional parameters and training complexity without clear benefits for our framework.

\subsubsection{Unified Training Objective}

The complete training objective combines all three constraints with appropriate weighting coefficients:

\begin{equation}
    \mathcal{L}_{\text{total}} = \alpha_{\text{ce}} \mathcal{L}_{\text{ce}} + \alpha_{\text{kl}} \mathcal{L}_{\text{kl}} + \alpha_{\text{sem}} \mathcal{L}_{\text{sem}},\label{equ:loss}
\end{equation}

\noindent where $\alpha_{\text{ce}}, \alpha_{\text{kl}}, \alpha_{\text{sem}} > 0$ are hyperparameters that balance the relative importance of answer correctness, prior alignment, and problem-thought semantic alignment, respectively. 

\subsection{Phase 3: Inference — Autoregressive Latent Reasoning with Termination}

A key aspect of autoregressive latent reasoning is deciding when to stop generating soft tokens and produce the final answer. In our framework, termination is trained explicitly through the cross-entropy loss on a special \texttt{</think>} token and the answer sequence.
Specifically, during training, each instance has a ground-truth explicit reasoning chain of variable length \(N\) (e.g., \(N=4\) in the example of Section~\ref{def:soft_token}). The model is trained to generate exactly \(N\) soft tokens (one per reasoning step), followed by a special \texttt{</think>} token, and finally the answer token sequence. Thus, the cross-entropy loss \(\mathcal{L}_{\text{ce}}\) supervises two aspects: (1) the emission of the \texttt{</think>} token after the \(N\)-th soft token, and (2) the correctness of the final answer.

At inference time, the model autoregressively generates soft tokens one by one. After each soft token, it decides whether to continue reasoning or to terminate by outputting \texttt{</think>}. In practice, we find that the model reliably produces \texttt{</think>} after generating a number of soft tokens close to the instance-specific step count seen during training. This instance-level variability is naturally learned because the training data contains problems with different \(N\), and the cross-entropy loss provides per-instance supervision.

\textbf{Training Stages vs. Inference Steps.} To avoid confusion, we explicitly distinguish three phases in our framework:

\begin{itemize}
    \item \textbf{Data Preprocessing (Phase 1)}: Rule-based priors are constructed offline from explicit reasoning chains. This is a one-time preprocessing step and is not part of model training or inference.
    \item \textbf{Training (Phase 2)}: The model is trained in a single stage using the joint loss defined in Eq.~\ref{equ:loss}. This is what we refer to as One-Step training—a single end-to-end training stage without cascaded pre-training or auxiliary models.
    \item \textbf{Inference (Phase 3)}: The model autoregressively generates soft tokens one by one, followed by the termination token and final answer. This is the standard autoregressive generation procedure for sequence models and should not be confused with the training-stage meaning of ``step''. The ``One-Model One-Step'' designation describes the training paradigm, not the inference generation process.
\end{itemize}

\section{Experiments}

\subsection{Experimental setup}

\textbf{Datasets.} We train and evaluate our method on GSM8k-Aug \citep{deng2023implicit}, an augmented variant of GSM8k \citep{cobbe2021training} for grade-school math, containing 385K training and 1K test samples. Notably, the reasoning chains in GSM8k-Aug are formalized as pure arithmetic expressions without natural language descriptions (e.g., \texttt{<<600*30/100=180>> <<600*10/100=60>>}), following the preprocessing of \cite{deng2023implicit}. This format naturally aligns with our step segmentation and token extraction pipeline. For out-of-domain evaluation, we use: (1) GSM-Hard \citep{gao2023pal} (1K test samples with larger numbers); (2) SVAMP \citep{patel2021nlp} (1K samples); and (3) MultiArith \citep{roy2015solving} (600 samples). Following \cite{tan2025think}, we report Accuracy (Acc.) and reasoning chain length (\# L), along with their ratio (Acc./\# L) to jointly measure performance and efficiency.

\textbf{Baseline methods.} We compare our method against several representative baselines: (1) \textbf{CoT} \citep{wei2022chain} is fine-tuned on complete reasoning chains and corresponding answers, performing token-level reasoning during inference before generating the final answer. (2) \textbf{iCoT} \citep{deng2024explicit} internalizes reasoning knowledge by progressively eliminating reasoning steps during training, enabling direct answer prediction without explicit reasoning at inference time. We also compare with three recent methods introduced in the related work: (3) \textbf{COCONUT} \citep{hao2025training}, (4) \textbf{CODI} \citep{shen2025codi}, (5) \textbf{CoLaR} \citep{tan2025think} and (6) \textbf{ReGuLaR} \citep{Wang2026ReGuLaR}.

\textbf{Implementation.} Unless otherwise specified, all experiments use a frozen Llama-3.2-1B-Instruct \citep{dubey2024llama} backbone with a trainable LoRA module \citep{hu2022lora}. Following COCONUT, all methods are initialized with weights from CoT-SFT to accelerate training. More details about experimental have been provided in Appendix \ref{app:experimental}.

\subsection{Mechanism-Level Comparison with Representative Latent CoT Methods}
\label{sec:mechanism_comparison}

To provide a clearer understanding of RuPLaR's position in the literature, we compare its core mechanisms with three representative approaches:

\textbf{COCONUT} \citep{hao2025training} operates by replacing discrete tokens with the final hidden state of the previous step, feeding it directly as input to the next step. This avoids discrete token generation for reasoning but suffers from spatial inconsistency between hidden states and token embeddings. RuPLaR instead operates in the vocabulary embedding space via soft token convex combinations, avoiding this mismatch.

\textbf{CoLaR} \citep{tan2025think} compresses reasoning via a learned latent head and reinforcement learning, enabling dynamic adjustment of compression ratio at inference time. RuPLaR's priors are constructed directly from the reasoning chain, providing more interpretable and stable supervision without RL.

\textbf{ReGuLaR} \citep{Wang2026ReGuLaR} formulates latent reasoning within a VAE framework, using rendered explicit reasoning chains as images to regularize the posterior distribution via visual-semantic representations. RuPLaR avoids the complexity of image rendering and visual feature extraction by using purely text-based, deterministic rule priors.

Compared with COCONUT's hidden-state approach, CoLaR's RL-augmented compression, and ReGuLaR's visually-regularized VAE framework, RuPLaR offers a distinct advantage through its \emph{One-Model One-Step} design: a single LLM handles both latent reasoning and answer generation in a single training stage, without cascaded pipelines, auxiliary models, or reinforcement learning. Combined with deterministic rule-based priors and selective KL alignment, this framework achieves stable, interpretable, and efficient latent reasoning with minimal system complexity.

\subsection{Main Results}

\begin{table*}[!ht]
    \centering
    \caption{\label{tab:main_ablation} Main results and ablation studies on four datasets. \#L denotes the average number of latent tokens generated during inference (autoregressive steps before termination). RuPLaR variants report mean $\pm$ std over 3 random seeds (777, 42, 1234).}
    \resizebox{\linewidth}{!}{
        \begin{tabular}{l | ccc|ccc|ccc|ccc|ccc }
            \toprule[2pt]
            \multirow{2}{*}{Model}
            & \multicolumn{3}{c|}{GSM8k-Aug}  
            & \multicolumn{3}{c|}{GSM-Hard} 
            & \multicolumn{3}{c|}{SVAMP} 
            & \multicolumn{3}{c|}{MultiArith} 
            & \multicolumn{3}{c}{Average} \\
            & Acc. & \# L & Acc./\# L & Acc. & \# L & Acc./\# L & Acc. & \# L & Acc./\# L & Acc. & \# L  & Acc./\# L & Acc. & \# L & Acc./\# L\\
            \midrule[1pt]
            CoT-SFT &49.4&25.6&1.93 &11.9&34.2&0.35 &59.8&12.1&4.94 &93.2&13.7&6.80 &53.6&21.4&2.50 \\
            iCoT    &19.8&0.00& -     &3.87&0.00& -      &36.4&0.00& -      &38.2&0.00& -     &24.6&0.00& - \\
            COCONUT &23.1&6.00&3.85   &5.49&6.00&0.92    &40.7&6.00&6.78    &41.1&6.00&6.85   &27.6&6.00&4.60 \\
            CODI    &13.3&6.00&2.22   &2.97&6.00&0.50    &21.7&6.00&3.62    &19.2&6.00&3.20   &14.3&6.00&2.38 \\
            CoLaR-5 &26.8&5.57&4.81   &5.87&6.53&0.90    &48.4&2.95&16.4    &86.4&3.21&26.9   &41.7&4.57&9.12 \\
            ReGuLaR &34.9&3.69&9.46   &8.27&4.12&2.01    &50.1&2.02&24.8    &89.8&2.04&39.1   &45.6&3.03&15.0 \\
            \midrule[1pt]
            RuPLaR-Temp     &44.24$\pm$0.14&2.97&14.9&9.86$\pm$0.00&2.78&3.55&\textbf{56.77$\pm$0.45}&1.50&\textbf{37.9}&89.82$\pm$0.02&2.04&44.0&50.17&2.32&21.6\\
            RuPLaR-Gumbel   &45.20$\pm$0.18&3.03&14.9&10.12$\pm$0.10&2.88&3.54&55.50$\pm$0.62&1.56&35.1&94.07$\pm$0.32&2.02&46.4&51.22&2.37&21.5\\
            RuPLaR-Mix      &\textbf{45.93$\pm$0.12}&3.00&\textbf{15.3}&\textbf{10.88$\pm$0.07}&2.87&\textbf{3.76}&53.93$\pm$0.32&1.59&34.2&\textbf{95.00$\pm$0.00}&2.03&\textbf{46.8}&\textbf{51.44}&2.37&\textbf{21.7}\\
            \midrule[2pt]
            \multicolumn{16}{l}{\textit{Rule-Based Prior Variants}} \\
            \quad Random Prior &34.1&2.81&12.1    &8.26&2.65&3.12   &48.6&1.42&34.2  &81.0&1.98&40.9  &43.0&2.22&19.4\\
            \quad Uniform Prior &35.1&3.03&11.6    &8.04&3.24&2.48   &47.3&1.56&30.3  &72.2&2.03&35.6  &40.7&2.47&16.5\\
            \quad Learned Prior &38.9&3.05&12.8    &9.78&2.88&3.40   &45.2&1.66&27.2  &82.3&2.07&39.8  &44.0&2.42&18.2\\
            \quad Result-Only Prior &22.7&2.66&8.53    &5.46&2.41&2.27   &40.6&1.64&24.8  &83.5&2.04&40.9  &38.1&2.19&17.4   \\
            \quad Operational-Only Prior &22.3&2.66&8.38    &5.69&2.41&2.36    &41.3&1.64&25.2   &83.2&2.04&40.8   &38.1&2.19&17.4  \\
            \midrule[1pt]
            \multicolumn{16}{l}{\textit{Prior Strength Variants}} \\
            \quad Weak (\(\alpha_{\text{kl}}=0.3\)) &40.9&2.83&14.5    &9.33&2.62&3.56   &52.5&1.43&36.7  &91.5&1.99&46.0  &48.6&2.22&21.9\\
            \quad Strong (\(\alpha_{\text{kl}}=2.0\)) &41.3&2.98&13.9    &9.47&2.87&3.30   &51.4&1.57&32.7  &89.2&2.01&44.4  &47.8&2.36&20.3\\
            \midrule[1pt]
            \multicolumn{16}{l}{\textit{Loss Component Variants}} \\
            \quad w/o KL-Focus  &39.6 &2.97 &13.3    &9.40&2.76&3.41   &48.9&1.60&30.6  &89.5&2.06&43.4  &46.9&2.35&20.0\\
            \quad w/o Problem-Thought  &39.1 &2.95 &13.3    &9.70&2.76&3.51   &51.0&1.56&32.7  &88.8&2.04&43.5  &47.3&2.33&20.3\\
            \bottomrule[2pt]
        \end{tabular}
    }
\end{table*}

We compare our model against six baselines, including CoT-SFT as a reference for explicit CoT reasoning. ``Temp'', ``Gumbel'', and ``Mix'' denote three variants for constructing rule-based priors (Section~\ref{sec:method}). CoLaR-5 compresses five reasoning tokens into one latent representation. Bold indicates best performance.

As shown in Table~\ref{tab:main_ablation}, RuPLaR consistently outperforms existing latent CoT methods. On GSM8k-Aug, RuPLaR-Mix achieves 45.93\% accuracy—an 11.03-point gain over ReGuLaR—with substantially fewer tokens. On GSM-Hard, it reaches 10.8\% accuracy, surpassing ReGuLaR by 2.53 points. On SVAMP, RuPLaR-Temp attains 56.8\% accuracy—6.7 points higher than ReGuLaR—using only 1.50 tokens. On MultiArith, RuPLaR-Mix achieves 95.0\% accuracy. Overall, RuPLaR-Mix leads with 51.44\% average accuracy, a 5.84\% absolute improvement over ReGuLaR, while using significantly fewer tokens. Compared to CoT-SFT, RuPLaR achieves slightly lower average accuracy (51.5\% vs. 53.6\%) but uses dramatically fewer tokens (2.37 vs. 21.4 tokens on average), representing a 9.0$\times$ reduction in generation length while largely preserving reasoning performance.

\textbf{Mixture Prior Superiority.} RuPLaR-Mix achieves the best overall performance among the three variants (51.44\% average accuracy, 21.7 Acc./\#L). Its mixture prior balances operational and result tokens via coefficient $\lambda$, providing more stable supervision than temperature-based or Gumbel approaches. This design captures both computational procedures and answer emphasis, yielding robust latent representations. RuPLaR-Gumbel achieves competitive results but slightly underperforms the mixture variant, suggesting that deterministic priors avoid sampling instability. RuPLaR-Temp excels on SVAMP with minimal token usage but slightly trails in overall accuracy.

\textbf{Efficiency-Effectiveness Trade-off.} RuPLaR achieves superior accuracy with minimal token usage. RuPLaR-Mix attains 21.7 Acc./\#L on average, substantially outperforming ReGuLaR, COCONUT, and CoLaR-5. This validates our core motivation: compressing multi-step reasoning with rule-based guidance enhances reasoning quality while reducing computational overhead. The fact that RuPLaR achieves this without cascaded models or inter-model dependencies underscores the effectiveness of our One-Model One-Step compression framework.

\subsection{Ablation Study}
\label{sec:ablation}

To comprehensively validate the effectiveness of our proposed Rule-Based Priors, we conduct extensive ablation experiments across all four mathematical reasoning datasets. Table~\ref{tab:main_ablation} presents the complete results, from which we draw several key insights.

\textbf{Rule-based Prior Variants.}
We first examine the contribution of our rule-based prior by comparing alternative configurations. Replacing rule-based scores with randomly assigned token importance (\textbf{Random Prior}) yields 43.0\% average accuracy—an 8.44-point drop from the full model's 51.44\%. Assigning equal importance to all tokens (\textbf{Uniform Prior}) performs even worse at 40.7\% (a 10.7-point decline), confirming that differentiated token importance is crucial. A data-driven prior based on token position statistics (\textbf{Learned Prior}) recovers some performance (44.0\%) but still lags 7.4 points behind, particularly on challenging datasets (GSM8k-Aug: 38.9\% vs. 45.93\%; MultiArith: 82.3\% vs. 95.0\%). To further isolate the contribution of each token category, we design \textbf{Result-Only Prior} and \textbf{Operational-Only Prior}, where the prior assigns probability exclusively to result tokens or operational tokens, respectively. Both variants achieve 38.1\% average accuracy—a 13.3-point drop from the full model—with substantial degradation on GSM8k-Aug (22.7\% and 22.3\%), indicating that neither token category alone provides sufficient supervision. The full model, combining both operational and result tokens via the mixture coefficient \(\lambda\), substantially outperforms all single-category priors, confirming that the complementary information from procedural semantics and outcome signals is essential for robust reasoning.

\textbf{Prior Strength Variants.}
We further examine the impact of the KL divergence weight $\alpha_{\text{kl}}$. A weaker prior ($\alpha_{\text{kl}}=0.3$) achieves 48.6\% accuracy with 2.22 steps, while a stronger prior ($\alpha_{\text{kl}}=2.0$) yields 47.8\% accuracy with 2.36 steps. The full model with $\alpha_{\text{kl}}=1.0$ strikes the optimal balance at 51.44\% accuracy and 2.37 steps. This consistent pattern across datasets demonstrates that moderate prior strength is optimal: too weak provides insufficient guidance, while too strong over-constrains the model and limits its flexibility.

\textbf{Loss Component Variants.}
Finally, we ablate two key loss components. Removing the focused KL-divergence (\textbf{w/o KL-Focus}) drops accuracy from 51.44\% to 46.9\%—a 4.54-point decrease—with consistent degradation, confirming its role in preserving semantic fidelity by filtering gradient noise. Removing the Problem-Thought Semantic Constraint (\textbf{w/o Problem-Thought}) yields 47.3\% accuracy—a 4.14-point drop—with declines across all datasets, demonstrating its function in grounding reasoning steps in the problem context. Together, these results validate that both components play complementary roles: KL-Focus ensures distributional correctness, while Problem-Thought maintains semantic coherence.

\subsection{Comparative Analysis with CoLaR Under K=2 Compression}

\begin{figure} 
	\centering
	\includegraphics[width=0.96\linewidth]{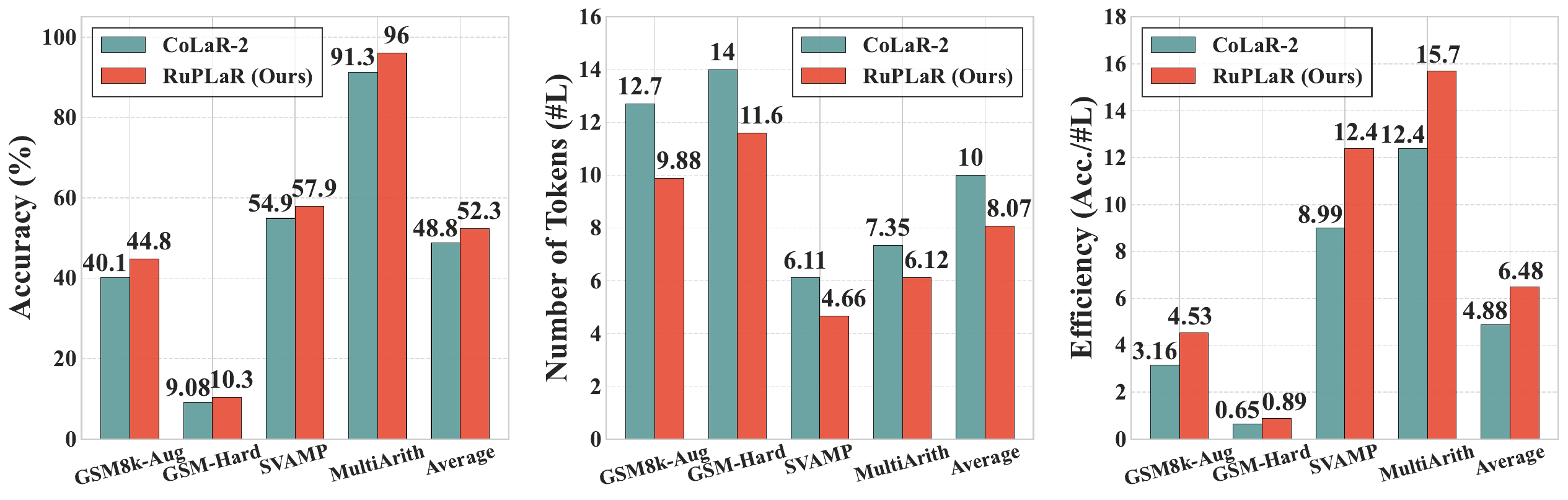}
	\caption{\label{fig:comparative_analysis_colar} Comparison with CoLaR-2 under identical compression ratio (k=2), showing accuracy, token usage, and efficiency across four datasets.}
\end{figure}

To ensure a fair comparison with CoLaR under the same compression setting, we evaluate our RuPLaR framework using k=2 compression, where every two reasoning tokens are compressed into a single latent token. Under this configuration, our method disables the Token Extraction component—operational tokens and result tokens are treated uniformly with equal probability assignment—enabling a direct and equitable comparison with CoLaR-2.

As shown in Figure~\ref{fig:comparative_analysis_colar}, RuPLaR consistently outperforms CoLaR-2 across all datasets. On average, RuPLaR achieves 52.3\% accuracy versus CoLaR-2's 48.8\%, a substantial 3.5\% absolute improvement, while using fewer tokens (8.07 vs. 10.0). This efficiency gain is further reflected in the Acc./\#L metric, where RuPLaR attains 6.48 compared to CoLaR-2's 4.88—a 32.8\% relative improvement.These results demonstrate that even without Token Extraction, RuPLaR's Rule-Based Priors and joint training objective enable superior latent representations under identical compression constraints. Moreover, they suggest that beyond simply increasing the number of soft tokens to boost performance, the Token Extraction component offers an additional orthogonal avenue for enhancement, which our full model leverages to achieve even greater gains.

\subsection{Scalability Analysis Across Model Sizes and Families}

\begin{table*}[!ht]
    \centering
    \caption{\label{tab:scalability} Scalability analysis across different model architectures and sizes. We evaluate LLaMA-3.2 (1B, 3B) and DeepSeek-R1-Distill-Qwen-1.5B as backbone LLMs.}
    \resizebox{\linewidth}{!}{
        \begin{tabular}{l | ccc|ccc|ccc|ccc|ccc }
            \toprule[2pt]
            \multirow{2}{*}{Backbone}
            & \multicolumn{3}{c|}{GSM8k-Aug}  
            & \multicolumn{3}{c|}{GSM-Hard} 
            & \multicolumn{3}{c|}{SVAMP} 
            & \multicolumn{3}{c|}{MultiArith} 
            & \multicolumn{3}{c}{Average} \\
            & Acc. & \# L & Acc./\# L & Acc. & \# L & Acc./\# L & Acc. & \# L & Acc./\# L & Acc. & \# L  & Acc./\# L & Acc. & \# L & Acc./\# L\\
            \midrule[1pt]
            \multicolumn{16}{l}{\textit{LLaMA Family}} \\
            LLaMA-1B     &46.0 &3.00 &15.3   &10.8&2.87&3.76    &54.3&1.59&34.2    &94.8&2.03&46.7    &51.4&2.37&21.7\\
            LLaMA-3B     &49.8&3.17&15.7     &11.2&3.09&3.62      &57.1&1.74&32.8       &96.7&2.05&47.1      &53.7&2.51&21.4  \\
            \midrule[1pt]
            \multicolumn{16}{l}{\textit{DeepSeek Family}} \\
            DS-1.5B      &42.7&3.07&13.9     &9.70&2.90&3.34      &53.3&1.60&33.3       &83.5&2.03&41.1      &47.3&2.40&19.7  \\
            \bottomrule[2pt]
        \end{tabular}
    }
\end{table*}

To assess the generalizability of our RuPLaR framework, we evaluate its performance across different backbone architectures and model scales. Specifically, we experiment with the LLaMA-3.2 family (1B and 3B parameters) and the DeepSeek-R1-Distill-Qwen-1.5B model. Table~\ref{tab:scalability} summarizes the results. All results in this subsection are obtained with a fixed random seed (seed=777) to isolate the effect of model architecture and scale, as the primary objective is cross-model comparison rather than statistical stability.

\textbf{Scaling Within the LLaMA Family.}
Increasing model capacity from 1B to 3B parameters consistently improves reasoning performance. The LLaMA-3B backbone achieves higher average accuracy (53.7\% vs. 51.4\%) and superior results on individual datasets, particularly on GSM8k-Aug (49.8\% vs. 46.0\%) and MultiArith (96.7\% vs. 94.8\%). This gain, however, comes with a modest increase in token usage (2.51 vs. 2.37), yielding comparable efficiency (21.4 vs. 21.7 Acc./\#L). These results indicate that larger models better leverage latent reasoning while maintaining similar token efficiency.

\textbf{Cross-Architecture Generalization.}
When applied to the DeepSeek-R1-Distill-Qwen-1.5B architecture, RuPLaR achieves 47.3\% average accuracy with 2.40 tokens, demonstrating strong cross-architecture transferability. Although this is lower than the LLaMA-1B counterpart (51.4\%), the performance remains competitive given the architectural differences. Notably, on SVAMP, DeepSeek achieves 53.3\% accuracy—approaching LLaMA-1B's 54.3\%—with reasonable efficiency (19.7 Acc./\#L). These findings suggest that RuPLaR's rule-based priors and joint training objective are architecture-agnostic and can be effectively applied across diverse model families.

\subsection{Analysis of Hyperparameters in Gumbel-Softmax Prior Construction}

\begin{figure}
	\centering
	\includegraphics[width=0.96\linewidth]{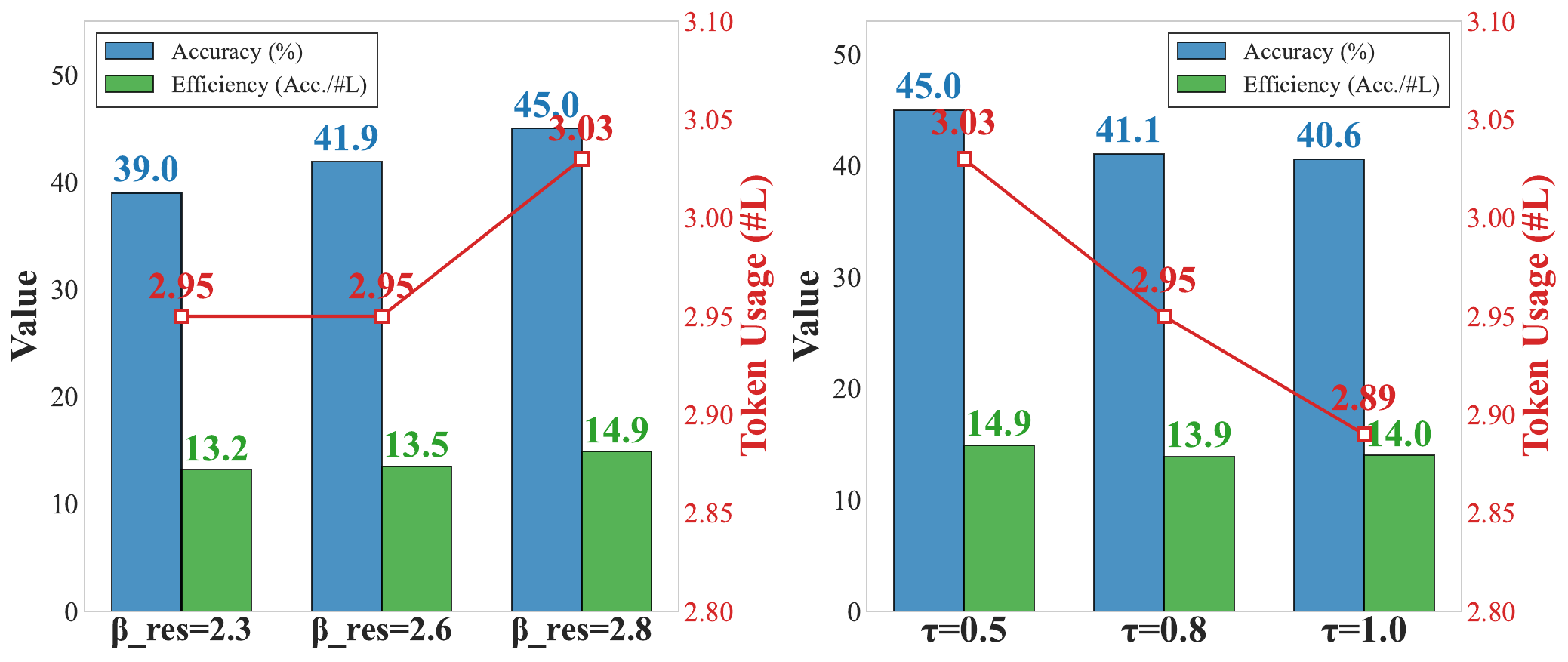}
	\caption{\label{fig:param_gumbel} Hyperparameter analysis on the GSM8k-Aug dataset, showing the impact of result token weight $\beta_{\text{res}}$ and temperature $\tau$ on accuracy, token usage, and efficiency.}
\end{figure}

We conduct a hyperparameter analysis for the Gumbel-Softmax prior construction on the GSM8k-Aug dataset, examining the impact of both the result token weight $\beta_{\text{res}}$ and temperature $\tau$ on model performance. As illustrated in Figure~\ref{fig:param_gumbel}, both hyperparameters significantly influence reasoning quality and efficiency. All results in this analysis are reported using a single seed (777) to focus on the trend of hyperparameter variation.

\textbf{Effect of Result Token Weight $\beta_{\text{res}}$.} 
Increasing $\beta_{\text{res}}$ from 2.3 to 2.8, while keeping $\beta_{\text{op}} = 2.0$ and $\tau = 0.5$ fixed, yields substantial performance gains. Accuracy rises from 39.0\% to 45.0\%—a notable 6.0\% absolute improvement—demonstrating that stronger emphasis on result tokens significantly enhances reasoning quality. Although token usage increases slightly from 2.95 to 3.03, the efficiency metric Acc./\#L improves markedly from 13.2 to 14.9, indicating that the performance gains outweigh the modest increase in token consumption. This confirms that appropriately prioritizing result tokens during prior construction plays a critical role in guiding the model toward more accurate reasoning.

\textbf{Effect of Temperature $\tau$.} 
With $\beta_{\text{res}} = 2.8$ and $\beta_{\text{op}} = 2.0$ fixed, temperature $\tau$ exhibits a clear inverse relationship with performance. The lowest temperature ($\tau = 0.5$) achieves the best results: 45.0\% accuracy and 14.9 efficiency. Increasing $\tau$ to 0.8 leads to a noticeable accuracy drop to 41.1\% (a 3.9-point decrease), with token usage slightly decreasing to 2.95. At $\tau = 1.0$, accuracy further declines to 40.6\% while token usage reduces to 2.89, though efficiency remains comparable at 14.0. This trend aligns with the theoretical properties of Gumbel-Softmax: as $\tau \to 0$, the distribution approaches a discrete categorical distribution, producing sharper, more deterministic samples that better preserve semantic focus. Conversely, higher temperatures introduce greater smoothing and stochasticity, which, while potentially beneficial for exploration during training, lead to less stable latent representations and diluted semantic signals.

\subsection{Analysis of Hyperparameters in Mix Prior Construction}

\begin{figure}
	\centering
	\includegraphics[width=0.96\linewidth]{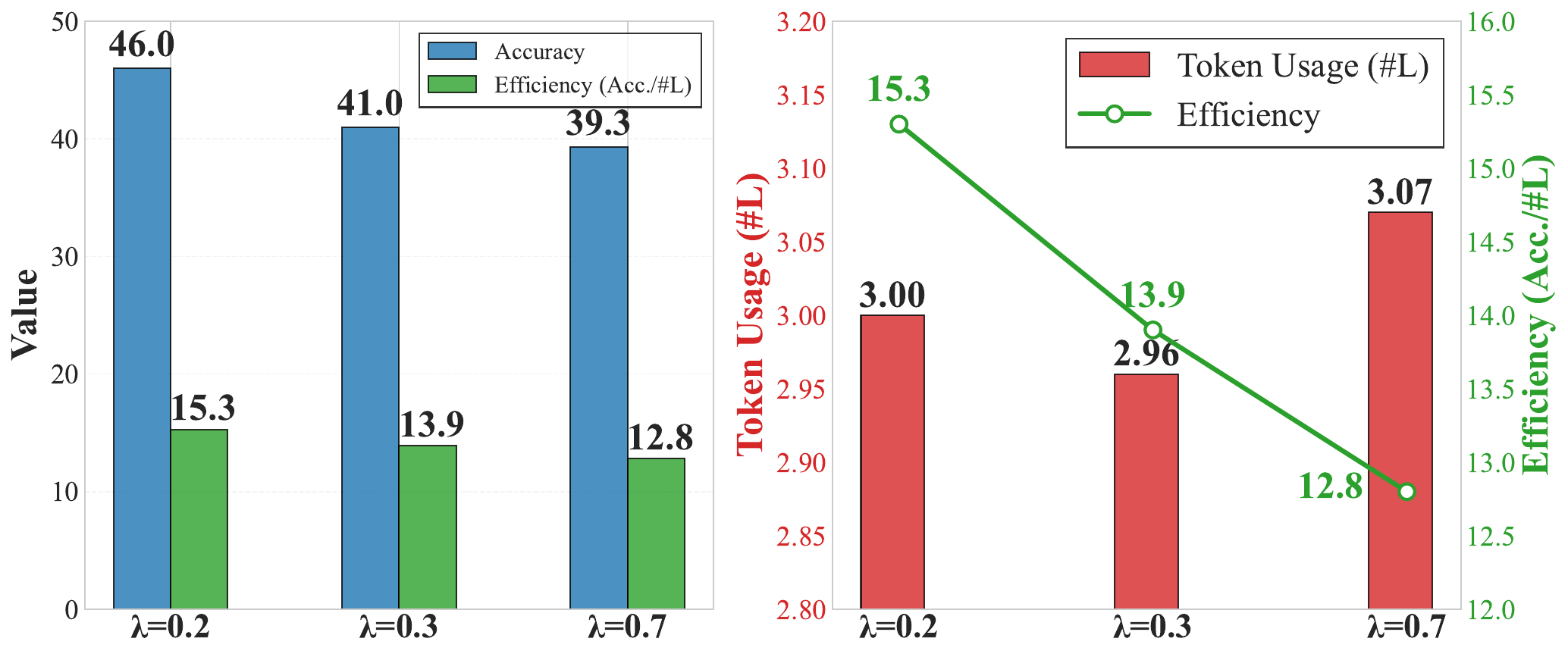}
	\caption{\label{fig:param_mix} Hyperparameter analysis on the GSM8k-Aug dataset, showing the impact of mixing coefficient $\lambda$ on accuracy, token usage, and efficiency.}
\end{figure}

We conduct a hyperparameter analysis for the mixture prior construction on the GSM8k-Aug dataset, focusing on the impact of the mixing coefficient $\lambda$, which controls the trade-off between emphasizing operational tokens (via uniform distribution) and the result token (via focused distribution). As shown in Figure~\ref{fig:param_mix}, we evaluate three settings: $\lambda = 0.2$, $\lambda = 0.3$, and $\lambda = 0.7$. Similarly, this analysis uses a fixed seed (seed=777) to isolate the effect of the mixing coefficient $\lambda$.

The results demonstrate that a lower $\lambda$, which assigns higher weight to the result token ($1-\lambda = 0.8$), yields the best performance. Specifically, with $\lambda = 0.2$, the model achieves 46.0\% accuracy using 3.00 tokens, resulting in the highest efficiency-accuracy ratio of 15.3. Increasing $\lambda$ to 0.3 leads to a noticeable accuracy drop to 41.0\% (a 5.0-point decrease), while token usage slightly decreases to 2.96, indicating that over-emphasizing operational tokens at the expense of the result token harms reasoning quality. At $\lambda = 0.7$, where the prior focuses predominantly on operational tokens, accuracy further declines to 39.3\% with increased token usage (3.07), yielding the lowest efficiency ratio of 12.8.

These findings underscore the critical role of the result token in guiding latent reasoning. While operational tokens capture the computational procedure, the result token anchors the final outcome, and a proper balance—favoring the result token—is essential for optimal performance. The mixture prior with $\lambda = 0.2$ achieves the best trade-off, validating our design choice and providing empirical guidance for hyperparameter selection in practice.

\subsection{Computational Efficiency}
\label{sec:efficiency}

We evaluate the computational efficiency of RuPLaR against baselines on GSM8k-Aug using a single NVIDIA A100 (40 GB) GPU with bfloat16 precision and FlashAttention 2 enabled. All models are evaluated under identical hardware and software conditions, with KV cache enabled during inference.

\subsubsection{Measurement Protocol}

All efficiency measurements are performed with a batch size of 1 and greedy decoding (temperature = 0) to ensure deterministic and reproducible results. All models use a maximum sequence length of 1024 tokens during inference, consistent with the training configuration, with KV cache enabled to avoid redundant computation. For latency measurements, we use \texttt{time.perf\_counter()} to measure the end-to-end inference time, including tokenization, the autoregressive generation loop (which may involve multiple forward passes to produce latent tokens and the final answer), and output decoding. The GPU is warmed up with 10 dummy runs before each measurement to stabilize performance. Latency values are averaged over 100 randomly sampled test instances, with standard deviations computed across three independent runs with different random seeds. Memory consumption is measured as peak GPU memory allocation via \texttt{torch.cuda.max\_memory\_allocated()}, with \texttt{torch.cuda.reset\_peak\_memory\_stats()} called before each run. All measurements are performed under identical conditions: models are loaded in evaluation mode with no optimizer states, and the same memory measurement API is used for all methods. Throughput is computed as the number of generated tokens per second, and the average number of generated tokens per sample is also reported for each method.

\begin{table*}[t]
    \centering
    \setlength{\tabcolsep}{5pt}
    \caption{\label{tab:efficiency} Computational efficiency comparison on GSM8k-Aug. Measured on a single NVIDIA A100 (40 GB) GPU with batch size 1 and greedy decoding. Latency values are means $\pm$ std over 100 samples; memory reported as peak allocation. Avg. Tokens reports the total number of generated tokens during inference, including latent tokens, stop tokens, and answer tokens.}
    \begin{tabular}{lccccc}
        \toprule[2pt]
        \textbf{Method} & \textbf{Latency (s)} & \textbf{Memory (GB)} & \textbf{Avg. Tokens} & \textbf{Throughput (tok/s)} & \textbf{Acc. (\%)} \\
        \midrule
        \multicolumn{6}{l}{\textit{Explicit CoT (Discrete Token Generation)}} \\
        \quad CoT-SFT & 0.770 $\pm$ 0.138 & 10.70 & 25.6 & \textbf{33.25} & 49.4 \\
        \midrule
        \multicolumn{6}{l}{\textit{Latent CoT (Latent Token Generation)}} \\
        \quad COCONUT   & 0.742 $\pm$ 0.131 & 7.99  & 6.00 & 8.09 & 23.1 \\
        \quad CoLaR-5   & 0.581 $\pm$ 0.183 & 10.11 & 5.57 & 9.59 & 26.8 \\
        \quad \textbf{RuPLaR} & \textbf{0.402 $\pm$ 0.054} & \textbf{2.37} & \textbf{3.04} & 7.56 & \textbf{45.93} \\
        \bottomrule[2pt]
    \end{tabular}
\end{table*}

\subsubsection{Efficiency Results}

As shown in Table~\ref{tab:efficiency}, RuPLaR achieves a mean latency of 0.402 seconds per sample, substantially outperforming all baselines: it is 47.8\% faster than CoT-SFT (0.770 s), 45.8\% faster than COCONUT (0.742 s), and 30.8\% faster than CoLaR-5 (0.581 s). This speedup stems from our One-Model One-Step compression framework, which eliminates cascaded training processes and inter-model dependencies. During inference, RuPLaR autoregressively generates a small number of latent tokens (3.04 on average) in a single model—without requiring auxiliary models or multi-stage pipelines—leading to substantially lower end-to-end latency compared to prior latent CoT approaches.

In terms of memory, RuPLaR requires only 2.37 GB of GPU memory during inference—a 77.8\% reduction over CoT-SFT (10.70 GB), a 70.3\% reduction over COCONUT (7.99 GB), and a 76.6\% reduction over CoLaR-5 (10.11 GB). To clarify the source of these differences, we note that CoLaR-5 incorporates an auxiliary Latent Head (16.79M parameters, as evidenced by its checkpoint size of 118,904 KB, 2.2$\times$ larger than COCONUT's 53,292 KB), which contributes to both the parameter footprint and intermediate state retention during generation. In contrast, RuPLaR achieves its memory savings through three factors: (i) no auxiliary components, as the focused KL-divergence loss directly guides the base LLM without additional parameters; (ii) shorter generation length (3.04 vs. 5.57 tokens for CoLaR-5), resulting in a smaller KV cache; and (iii) no intermediate state retention across autoregressive steps. The reported memory values reflect peak incremental memory during generation, with all methods using the same measurement protocol.

\textbf{Token-Length Compression.}
Table~\ref{tab:efficiency} reveals a key difference in generation length. CoT-SFT generates 25.6 discrete tokens per sample on average, while latent CoT methods compress reasoning into compact tokens: COCONUT and CoLaR-5 use 6.00 and 5.57 tokens respectively, and RuPLaR further reduces this to just \textbf{3.04 tokens}—an 8.4$\times$ compression ratio over CoT-SFT. This token reduction directly enables RuPLaR's latency gains. Notably, RuPLaR achieves competitive per-token throughput (7.56 tok/s) compared to COCONUT (8.09) and CoLaR-5 (9.59), while generating significantly fewer tokens per instance. This indicates that RuPLaR's speedup primarily comes from its compact reasoning representation rather than faster per-token generation.

Crucially, RuPLaR achieves these efficiency gains without sacrificing accuracy, attaining 45.93\%—closely approaching CoT-SFT (49.4\%) while far exceeding COCONUT (23.1\%) and CoLaR-5 (26.8\%). This represents an 11.03 percentage-point improvement over ReGuLaR on GSM8k-Aug, validating our core motivation: compressing multi-step reasoning with rule-based guidance enhances reasoning quality while dramatically reducing computational overhead, making RuPLaR particularly suitable for resource-constrained deployment.

\subsection{Analysis of Latent Reasoning Dynamics}
\label{sec:analysis}

\begin{figure*}[!ht]
    \centering
    \begin{tabular}{cc}
        \begin{minipage}{0.48\textwidth}
            \centering
            \includegraphics[width=\linewidth]{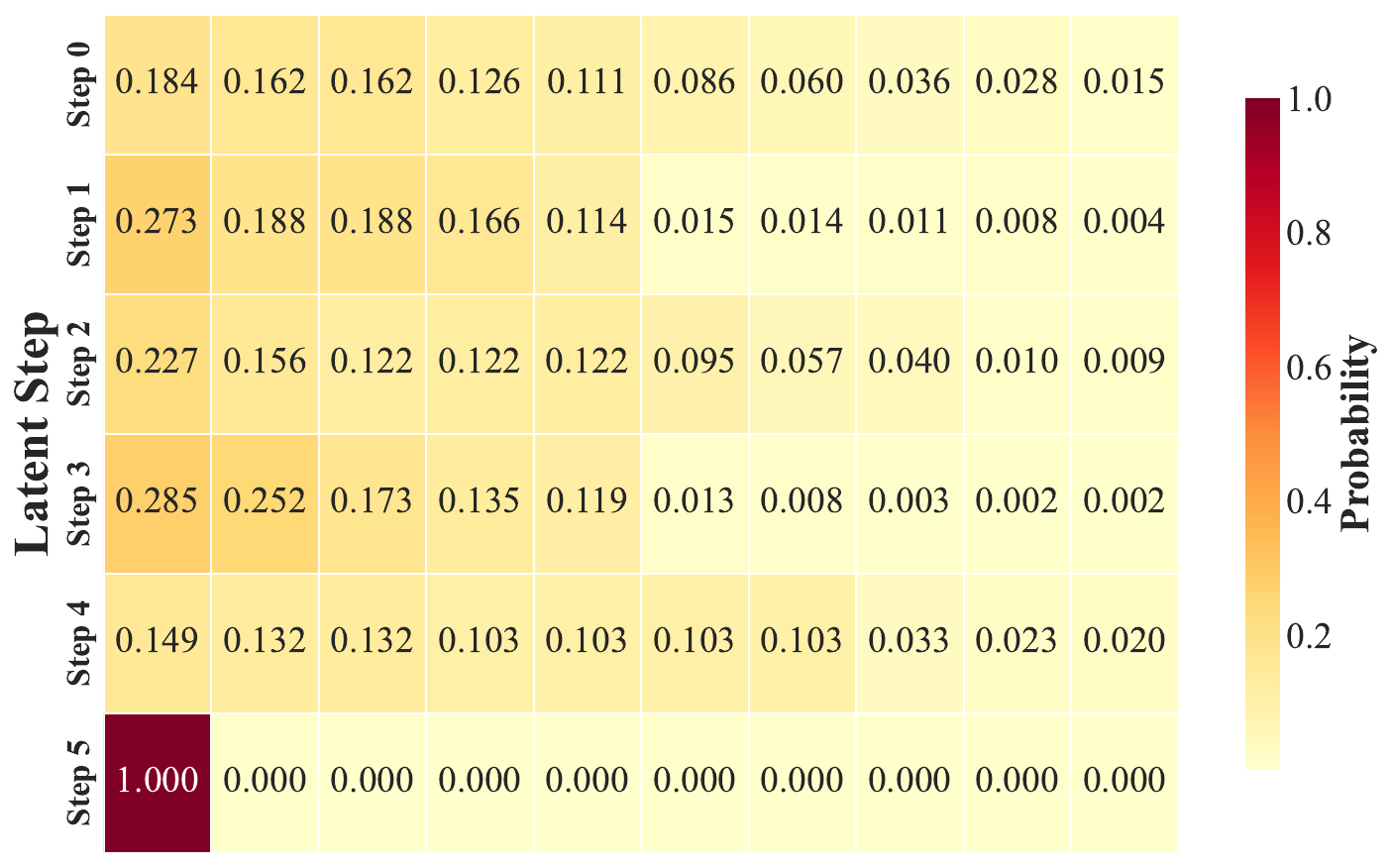}
            \subcaption{Top-10 Token Probabilities.}
            \label{fig:heatmap}
        \end{minipage} &
        \begin{minipage}{0.48\textwidth}
            \centering
            \includegraphics[width=\linewidth]{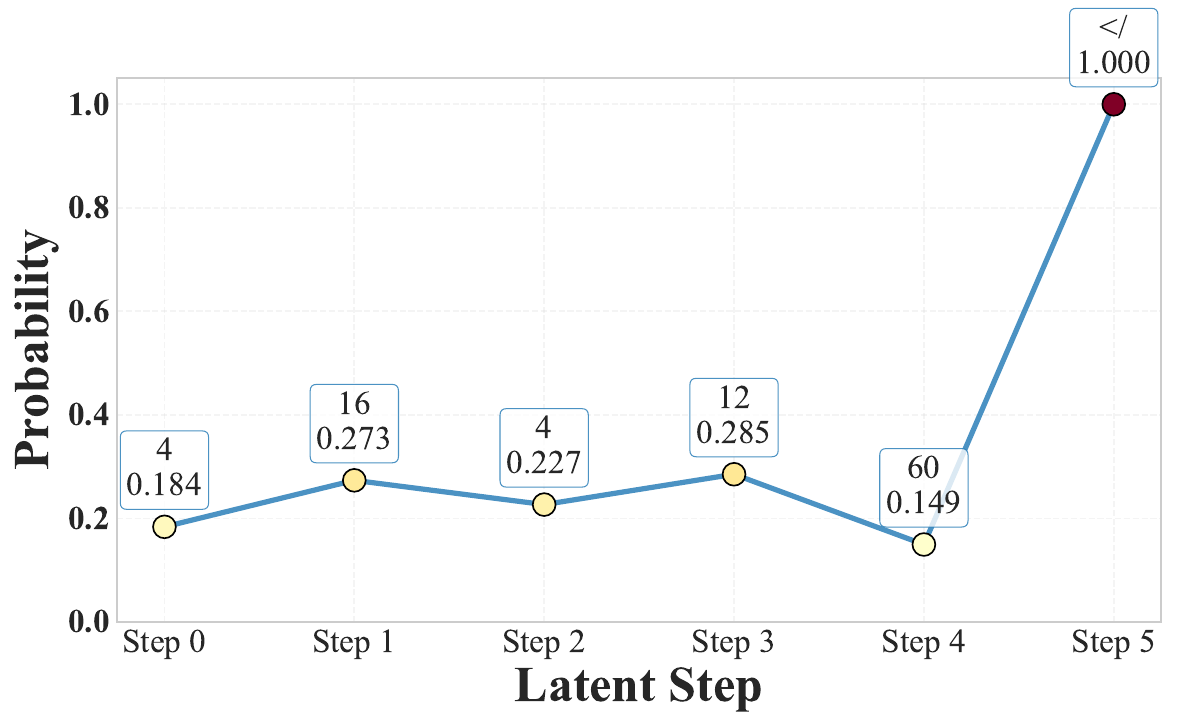}
            \subcaption{Top-1 Token Probability Trend.}
            \label{fig:top1_trend}
        \end{minipage} \\
        \begin{minipage}{0.48\textwidth}
            \centering
            \includegraphics[width=\linewidth]{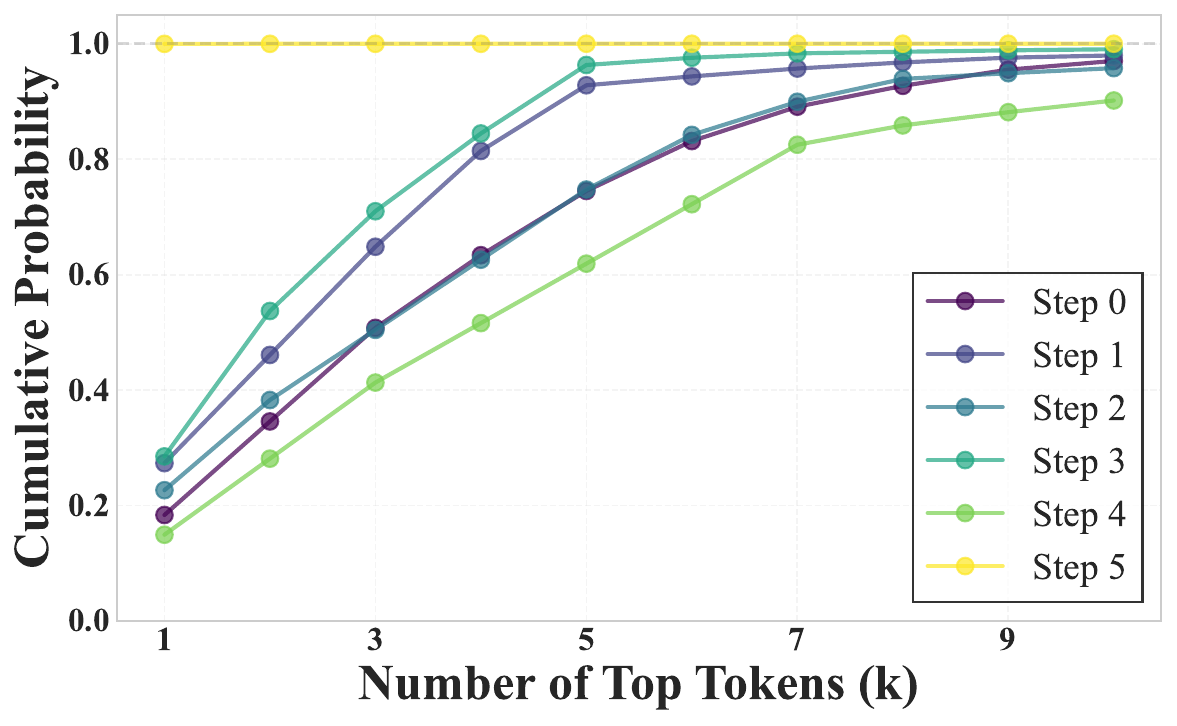}
            \subcaption{Cumulative Probability.}
            \label{fig:cumulative}
        \end{minipage} &
        \begin{minipage}{0.48\textwidth}
            \centering
            \includegraphics[width=\linewidth]{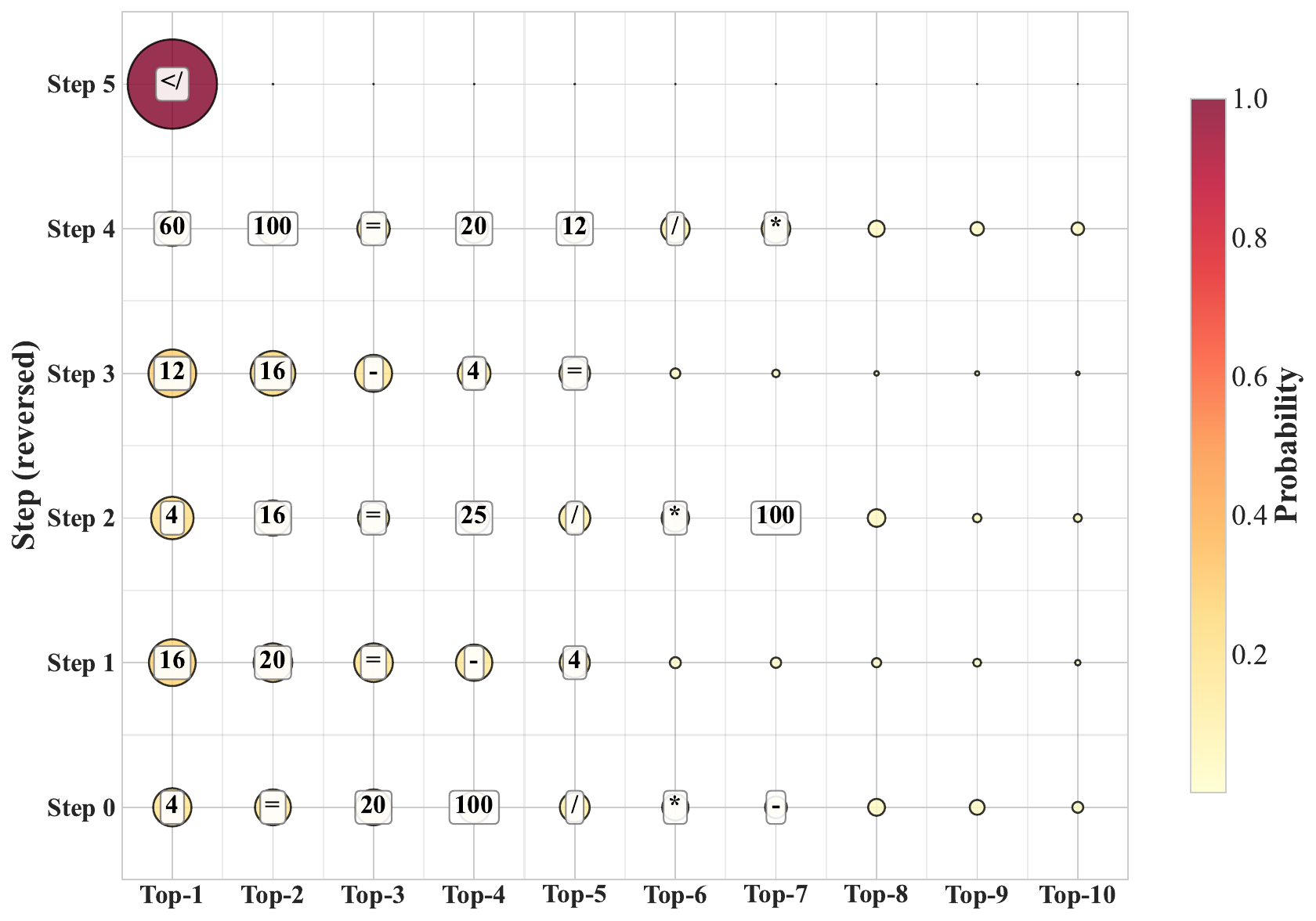}
            \subcaption{Token Distribution (Bubble Size = Probability).}
            \label{fig:bubble}
        \end{minipage}
    \end{tabular}
    \caption{Analysis of latent reasoning dynamics. The visualizations illustrate (a) the probability distribution of top-10 tokens across latent steps, (b) the trend of top-1 token probabilities with corresponding token labels, (c) cumulative probabilities of top-k tokens at each step, and (d) a bubble chart where bubble size and color intensity represent token probability magnitude.}
    \label{fig:latent_visualization}
\end{figure*}

To gain insight into the latent reasoning process, we visualize the probability distributions of soft tokens for a representative example from GSM8k-Aug, where the model generates six soft tokens before producing the final answer. The problem is: \emph{``In a dance class of 20 students, 20\% enrolled in contemporary, 25\% of remaining in jazz, and rest in hip-hop. What percentage are in hip-hop?''} The explicit reasoning chain comprises five sequential calculations: step 1 computes contemporary enrollment (\(20 \times 20/100 = 4\)), step 2 calculates remaining students (\(20 - 4 = 16\)), step 3 determines jazz enrollment (\(16 \times 25/100 = 4\)), step 4 obtains hip-hop enrollment (\(16 - 4 = 12\)), and step 5 derives the final percentage (\(12/20 \times 100 = 60\)).

Figure~\ref{fig:latent_visualization} visualizes the corresponding latent reasoning dynamics. The token probability heatmap (a) shows that while early steps exhibit distributed probabilities reflecting exploratory behavior, step 5 assigns near-certain probability (1.0) to the termination token \texttt{</think>}, indicating confident reasoning completion. This explore-then-converge pattern is further evidenced by the top-1 probability trend (b), where confidence remains moderate (0.18–0.29) during the first four steps before sharply rising to 0.99 at step 5. The cumulative probability analysis (c) reveals that early steps concentrate probability mass on few tokens, demonstrating focused attention, while the near-vertical line at step 5 confirms deterministic termination.

Most notably, the token distribution bubble chart (d) reveals that latent tokens closely mirror the explicit arithmetic progression: step 0 captures the first operation (``4'', ``20'', ``100''), step 1 encodes the intermediate result (``16''), step 2 reflects the next calculation (``4'', ``25''), step 3 represents the subsequent outcome (``12'', ``16''), step 4 encodes the final answer (``60''), and step 5 terminates with \texttt{</think>}. This alignment between latent representations and explicit computational steps demonstrates that RuPLaR successfully distills multi-step arithmetic reasoning into compact latent tokens while preserving logical structure and termination behavior, validating the effectiveness of our Rule-Based Priors and joint training objective.

\subsection{Batch Statistics Analysis}
\label{sec:batch_statistics}

\begin{figure*}[!ht]
    \centering
    \begin{tabular}{cc}
        \begin{minipage}{0.48\textwidth}
            \centering
            \includegraphics[width=\linewidth]{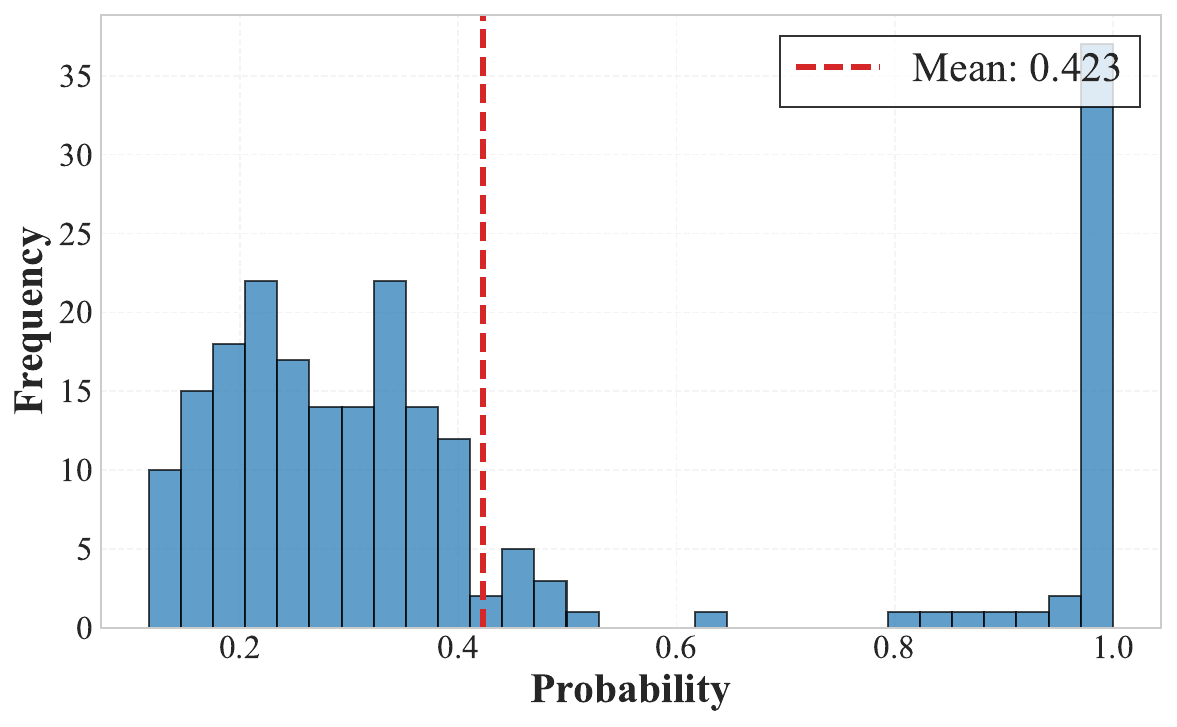}
            \subcaption{Top-1 Probability Distribution.}
            \label{fig:batch_top1}
        \end{minipage} &
        \begin{minipage}{0.48\textwidth}
            \centering
            \includegraphics[width=\linewidth]{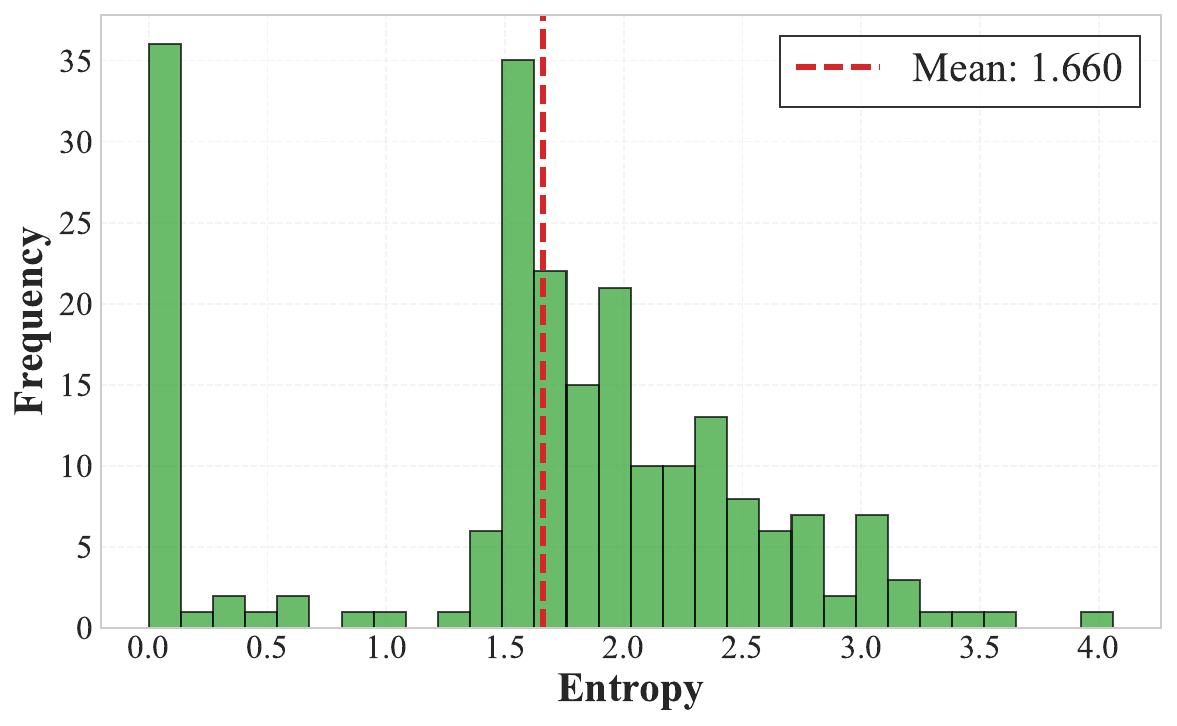}
            \subcaption{Entropy Distribution.}
            \label{fig:batch_entropy}
        \end{minipage} \\
        \begin{minipage}{0.48\textwidth}
            \centering
            \includegraphics[width=\linewidth]{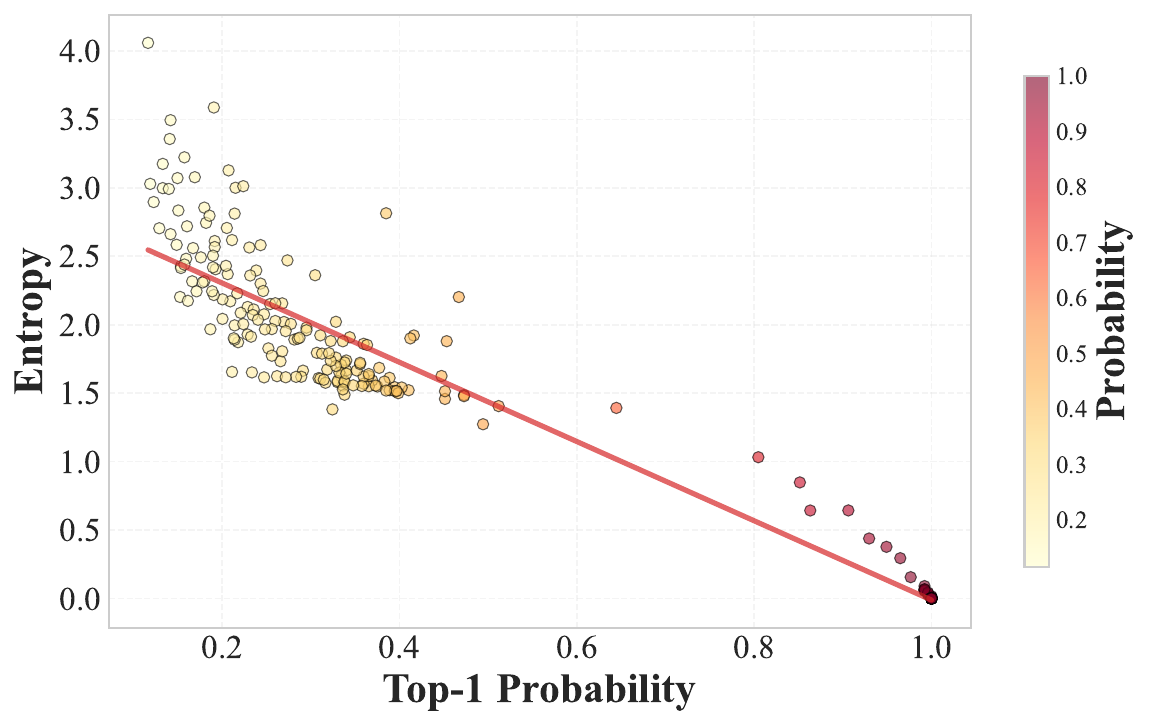}
            \subcaption{Top-1 Probability vs. Entropy.}
            \label{fig:batch_scatter}
        \end{minipage} &
        \begin{minipage}{0.48\textwidth}
            \centering
            \includegraphics[width=\linewidth]{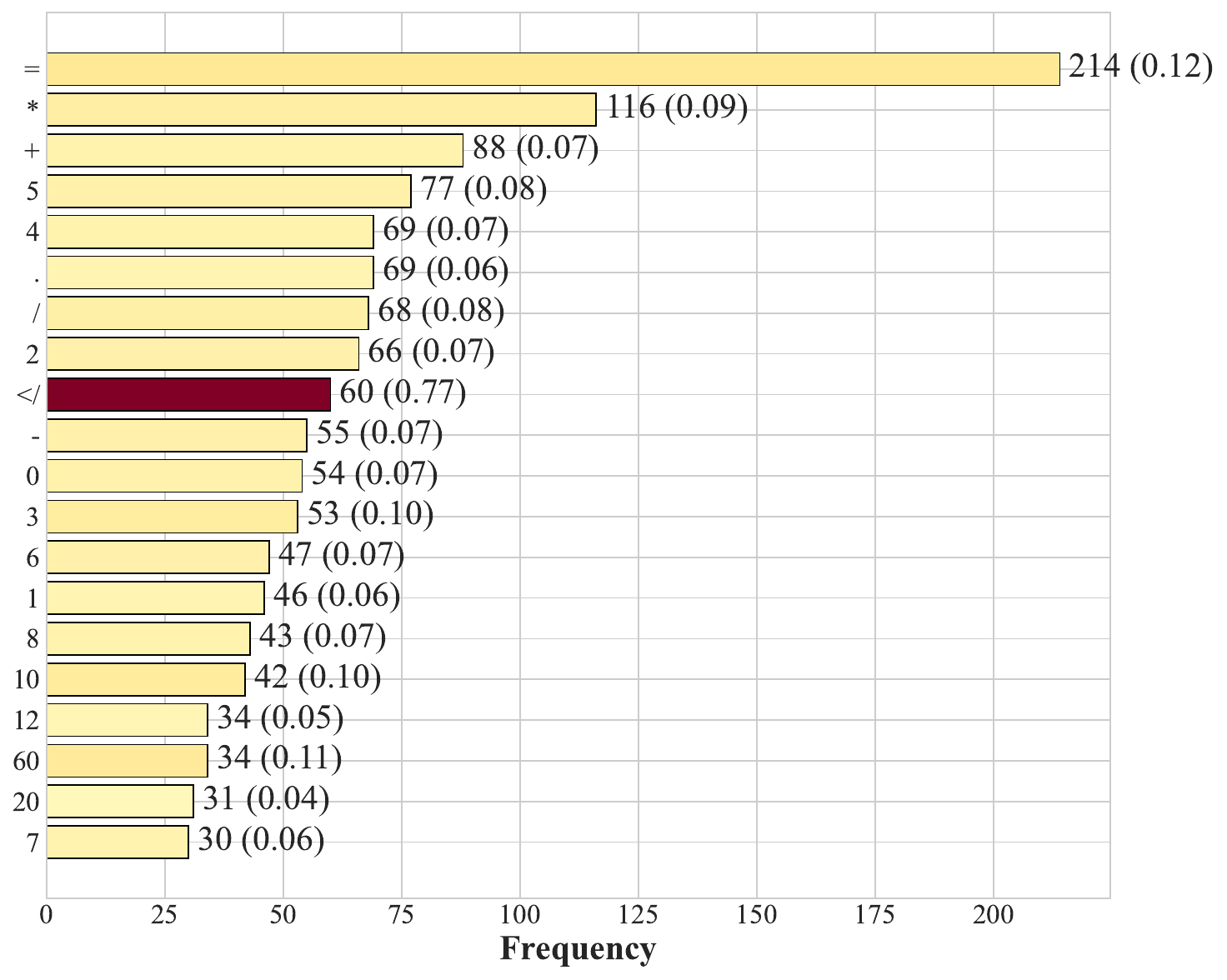}
            \subcaption{Top-20 Most Frequent Tokens.}
            \label{fig:batch_freq}
        \end{minipage}
    \end{tabular}
    \caption{Batch statistics analysis across 50 GSM8k-Aug instances. The visualizations show (a) distribution of top-1 token probabilities, (b) distribution of entropy values, (c) relationship between top-1 probability and entropy, and (d) frequency of the most common tokens in latent representations.}
    \label{fig:batch_statistics}
\end{figure*}

To understand the aggregate behavior of our latent reasoning model, we analyze statistics from 50 randomly sampled GSM8k-Aug instances. Figure~\ref{fig:batch_statistics} reveals key patterns in the latent reasoning process. The top-1 probability distribution (Figure~\ref{fig:batch_top1}) exhibits a clear bimodal pattern (mean 0.423), with peaks in the moderate range (0.2–0.4) and near 1.0, indicating distinct exploration and termination phases. This pattern is mirrored in the entropy distribution (Figure~\ref{fig:batch_entropy}, mean 1.660), which shows corresponding peaks at near-zero and moderate values. The strong negative correlation between top-1 probability and entropy (Figure~\ref{fig:batch_scatter}) further validates this two-phase dynamic, with points clustering in two distinct regions corresponding to exploratory and terminating steps.

Token frequency analysis (Figure~\ref{fig:batch_freq}) identifies three dominant categories aligned with our Rule-Based Priors design: arithmetic operators (\texttt{=}: 214, \texttt{*}: 116, \texttt{/}: 68, \texttt{-}: 55), numerical values (\texttt{4}: 69, \texttt{2}: 66, \texttt{3}: 53, \texttt{6}: 47, \texttt{20}: 31), and the termination token \texttt{</think>} (60). The prevalence of \texttt{=} underscores its central role in mathematical expressions, while the diversity of numbers reflects the variety of arithmetic operations encountered in the dataset. Together, these statistics confirm that RuPLaR learns a structured latent reasoning process characterized by clear exploration-termination dynamics and strong alignment with underlying arithmetic semantics.

\subsection{Generalization to Symbolic Temporal Reasoning: Date Understanding}

To provide an initial indication of whether our method can extend beyond mathematical reasoning, we conduct a preliminary experiment on the \textbf{Date Understanding (DU)} dataset \citep{srivastava2023beyond}. We emphasize that this experiment is limited in scale (369 samples) and is intended as a proof-of-concept rather than a comprehensive demonstration of cross-domain generalizability. The DU dataset evaluates the ability to perform date arithmetic and logical reasoning from natural language descriptions. The reasoning steps involve unit conversions (day, month, year), leap year handling, and relative date calculations (e.g., ``three days ago''). Although the original problems are expressed in natural language, each step can be naturally abstracted as a symbolic arithmetic expression (e.g., `MM/DD/YYYY + 1 day = MM/DD/YYYY').

\textbf{Data Conversion.}
The original Date Understanding dataset provides each sample with a natural language question (`input') and multiple-choice answers (`target\_scores'), where the correct answer is marked with a score of 1. To convert this dataset into a format compatible with our framework (triplet of question, reasoning chain, and answer), we employ DeepSeek, a large language model, to automate the conversion process. DeepSeek is prompted to perform three tasks for each sample: (1) identify and extract the correct answer from the multiple-choice options; (2) parse the natural language temporal relations (e.g., ``yesterday'', ``three days later'') and rewrite them as symbolic arithmetic steps enclosed by `<< >>'; and (3) enclose the final answer within `\texttt{\textbackslash boxed{}}'. After processing all samples, we obtain 369 valid instances, which are then randomly split into 295 training and 74 testing examples.

\begin{figure}
    \centering
    \begin{minipage}{0.23\textwidth}
        \centering
        \includegraphics[width=\linewidth]{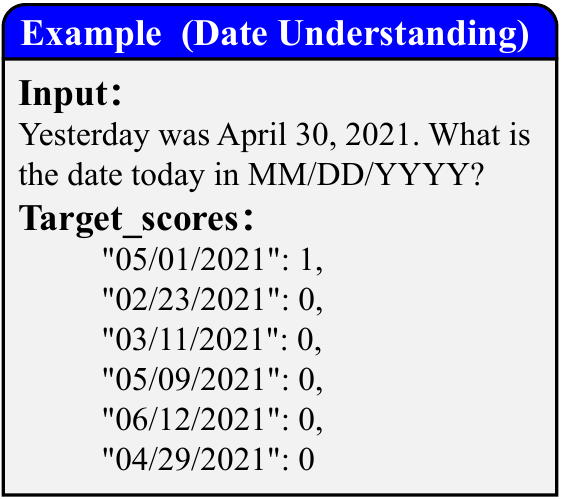}
        \label{fig:example_Du}
    \end{minipage}
    \begin{minipage}{0.23\textwidth}
        \centering
        \includegraphics[width=\linewidth]{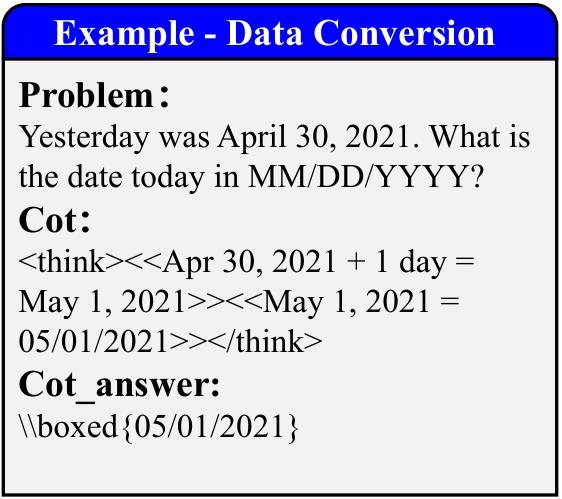}
        \label{fig:Du_DataConversion}
    \end{minipage}
    \caption{The data conversion process for the Date Understanding dataset, with examples before and after conversion.}
    \label{fig:data_conversion}
\end{figure}

\textbf{Results.}
We compare RuPLaR against SoftCoT \citep{xu-etal-2025-softcot} as a baseline on the Date Understanding dataset. SoftCoT directly outputs the final answer without generating explicit reasoning tokens (hence its average token length is marked as ``--'' in Table~\ref{tab:date_results}), serving as a reference for direct answer prediction. Our method achieves an accuracy of \textbf{82.4\%} on the Date Understanding test set, substantially outperforming SoftCoT (59.0\%), while using only 1.0 latent token on average per instance. The high accuracy with minimal reasoning steps suggests that most date questions can be resolved with a single symbolic operation, providing preliminary evidence that our symbolic abstraction may be applicable to certain non-mathematical tasks. However, we acknowledge that this experiment is limited in scale (369 samples) and only includes one baseline comparison. Larger-scale evaluations on more diverse reasoning domains—with comparisons to additional baselines—are necessary to firmly establish the generalizability of the framework.

\begin{table}[ht]
    \centering
    \small
    \caption{\label{tab:date_results} Results on the Date Understanding dataset. Avg. Tokens denotes the average number of generated tokens per instance; ``--'' indicates the model directly outputs the answer without explicit reasoning generation.}
    \begin{tabular}{lcc}
        \toprule
        Model & Accuracy & Avg. Tokens \\
        \midrule
        SoftCoT & 59.0\% & -- \\
        \textbf{RuPLaR} & \textbf{82.4\%} & 1.0 \\
        \bottomrule
    \end{tabular}
\end{table}

\section{Conclusion}

We introduce One-Model One-Step, a compression framework for RuPLaR that trains a single LLM to generate latent reasoning tokens in a single training stage guided by Rule-Based Priors, eliminating cascaded processes and inter-model dependencies. A joint loss function combines answer consistency, focused KL-divergence for selective token alignment, and a problem-thought constraint to ground reasoning in the question context. Experiments on four mathematical benchmarks show RuPLaR achieves 45.93\% accuracy on GSM8k-Aug, outperforming ReGuLaR by 11.03 points with fewer tokens. Collectively, these findings establish RuPLaR as an effective and efficient paradigm for latent reasoning, offering a robust foundation for future research in this direction.

\section*{Limitations}
\label{Limitations}

Despite the promising results achieved by RuPLaR, several limitations warrant discussion. 

First, the prior construction methods require careful hyperparameter tuning ($\beta_{\text{res}}$, $\beta_{\text{op}}$, $\lambda$, $\tau$), as suboptimal settings cause significant performance degradation. 

Second, while we provide a preliminary experiment on Date Understanding, the empirical evaluation in this paper is predominantly centered on mathematical reasoning. Although the framework is conceptually designed to be broadly applicable, comprehensive validation on other reasoning categories—such as logical deduction, commonsense reasoning, or causal inference—has not yet been conducted. We therefore encourage readers to interpret our generalizability claims with appropriate caution. 

Third, the effectiveness of our rule-based prior construction depends on the availability of explicit, segmentable symbolic structures in the reasoning chain. In domains such as mathematical arithmetic, symbolic temporal reasoning, or program execution, reasoning steps naturally decompose into discrete computational units with clear input-output boundaries, making our segmentation and token extraction strategies directly applicable. However, for reasoning domains where the underlying reasoning process is less structured—such as commonsense reasoning (e.g., `Why did the person take an umbrella?'), causal inference, or open-ended narrative reasoning—the lack of explicit symbolic expressions poses a fundamental challenge. In such cases, identifying `operational tokens' and `result tokens' is not well-defined, and the prior construction pipeline would require either (i) human annotation of step boundaries and key concepts, which is expensive and subjective, or (ii) alternative heuristics for extracting semantically meaningful units from free-form text, which may introduce noise or fail to capture the underlying reasoning logic. We therefore view our current framework as most suitable for structured reasoning tasks where symbolic or semi-symbolic intermediates are naturally available, and leave the extension to less structured domains as an important direction for future work.

\section*{Acknowledgments}

This work was supported by the National Natural Science Foundation of China (Grant Nos. 62537001 and 62276179) and the Project Funded by the Priority Academic Program Development of Jiangsu Higher Education Institutions.



\bibliographystyle{cas-model2-names}

\bibliography{cas-refs}

\appendix

\section{More Experimental Details}
\label{app:experimental}

\subsection{Datasets}
\label{app:datasets}

Following prior work \citep{tan2025think}, we train and evaluate our method primarily on the GSM8K-Aug dataset \citep{deng2023implicit}, and further assess its generalization on three out-of-domain mathematical reasoning datasets: GSM-Hard \citep{gao2023pal}, SVAMP \citep{patel2021nlp}, and MultiArith \citep{roy2015solving}. Detailed descriptions of these datasets are provided below:

\begin{itemize}
    \item \textbf{GSM8K-Aug}: An augmented version of the original GSM8K dataset \citep{cobbe2021training}, constructed by prompting GPT-4 \citep{achiam2023gpt} to expand the training set to 385K samples. Notably, it removes natural language descriptions from the reasoning process, formalizing each reasoning step as a mathematical expression.
    \item \textbf{GSM-Hard}: A harder variant of GSM8K, created by replacing numbers in the original test set with larger, less common values. This dataset serves as an out-of-domain benchmark in our experiments.
    \item \textbf{SVAMP}: Consists of 1,000 elementary-level math word problems designed to test model robustness on fundamental mathematical tasks. It is used as an out-of-domain dataset in our evaluation.
    \item \textbf{MultiArith}: Contains 600 multi-step arithmetic problems aimed at assessing a model's capacity for multi-step reasoning. This dataset also serves as an out-of-domain benchmark.
\end{itemize}

\subsection{Implementation Details}

We fine-tune all models using LoRA \citep{hu2022lora} with rank \(r=32\) and scaling factor \(\alpha=64\), applying adaptations to all linear layers in the attention and feed-forward modules---including \texttt{q\_proj}, \texttt{k\_proj}, \texttt{v\_proj}, \texttt{o\_proj}, \texttt{gate\_proj}, \texttt{up\_proj}, \texttt{down\_proj}---as well as the token embedding layer \texttt{embed\_tokens}. 

All experiments are conducted on a single NVIDIA A100 (40 GB) GPU using DeepSpeed ZeRO-1 \citep{rasley2020deepspeed} for efficient memory management. Training uses the AdamW optimizer \citep{loshchilov2017decoupled} with a learning rate of \(3 \times 10^{-4}\), warmup ratio of 0.05, weight decay of 0.01, and cosine learning rate scheduler. We train for up to 10 epochs and select the checkpoint with the highest validation accuracy as the final model. The per-device training batch size is 32 with gradient accumulation steps of 8, yielding an effective batch size of 256. All models are trained in bfloat16 precision. The maximum sequence length is set to 1024 tokens. FlashAttention 2 \citep{diao2023flashattention} is enabled for both training and inference. The random seed is fixed to 777 for reproducibility. All baseline methods are trained under the identical configuration described above, ensuring a fair comparison.

The software environment consists of Python 3.12.3, PyTorch 2.5.1, Transformers 4.36.0, DeepSpeed 0.17.0, PEFT 0.15.2, FlashAttention 2.7.3, and CUDA 12.1. Additional dependencies include datasets 3.6.0, scikit-learn 1.7.0, pylatexenc 2.10, latex2sympy2 1.9.1, and word2number 1.1.

Training batch sizes are set to 32, 10, and 24 for Llama-3.2-1B-Instruct, Llama-3.2-3B-Instruct, and DeepSeek-R1-Distill-Qwen-1.5B, respectively (with gradient accumulation adjusted to maintain comparable effective batch sizes). For the loss functions, we set \(\beta_{\text{res}} = 2.8\), \(\beta_{\text{op}} = 2.0\), and \(\tau = 0.5\) in Equation~\ref{equ:Temp}; \(\delta = 1 \times 10^{-2}\) in Equation~\ref{equ:kl}; and \(\alpha_{\text{ce}} = \alpha_{\text{kl}} = \alpha_{\text{sem}} = 1\) in Equation~\ref{equ:loss}.

\textbf{Focused KL-Divergence.}
For the focused KL-divergence loss (Eq.~\ref{equ:kl}), the threshold is set to \(\delta = 1 \times 10^{-2}\). The teacher distribution \(p_{\text{prior}}\) is first normalized over the full vocabulary; the mask is then applied to filter out tokens with probability below \(\delta\). During loss computation, the gradient only propagates through the selected tokens, effectively reducing the computational cost of the gradient step from O(|V|) to O(k) per alignment position. The forward pass still computes logits over the full vocabulary, as this is inherent to the model's output head. This selective masking is implemented via an element-wise multiplication of the per-token KL contribution with the mask tensor, ensuring that only the top contributing tokens influence the loss and gradient. Empirically, we find that \(k\) is typically in the range of 5--20 tokens for the GSM8k-Aug dataset, compared to \(|V| \approx 128,000\), yielding a substantial reduction in gradient computation cost.

\section{Rule-Based Priors Construction Pipeline}
\label{app:rule_based_priors}

\begin{algorithm*}[t]
\caption{Rule-Based Priors Construction Pipeline}
\label{alg:rule_based_priors}
\begin{algorithmic}[1]
\Require 
    \begin{itemize}
        \itemsep0em 
        \item Explicit reasoning chain $\mathbf{t}_r = (t_r^1, \dots, t_r^{L_r})$
        \item Vocabulary $V$, token embeddings $\{e_j\}_{j \in V}$
        \item Hyperparameters: $\tau$, $\beta_{\text{op}}$, $\beta_{\text{res}}$, $\lambda$, $k$, $\delta$
        \item Prior construction method $M \in \{\text{Temp}, \text{Gumbel}, \text{Mix}\}$
    \end{itemize}
\Ensure Sequence of prior distributions $\{\pi_i\}_{i=1}^N$ over $V$

\State \textbf{Step 1: Step Segmentation}
\State Parse $\mathbf{t}_r$ into computational steps $\{r_i\}_{i=1}^N$ using pattern-based delimiters (e.g., \texttt{<<...>>})

\For{$i \gets 1$ to $N$}
    \State \textbf{Step 2: Token Extraction}
    \State Identify operational token set $\mathcal{K}_i \subset V$ from $r_i$
    \State Identify result token $j_{\text{res}}^i \in V$
    
    \State \textbf{Step 3: Logit Initialization}
    \State Initialize logit vector $\bm{\ell} \in \mathbb{R}^{|V|}$ with $\ell_j \gets -\infty$ for all $j \in V$
    \For{each $j \in \mathcal{K}_i$}
        \State $\ell_j \gets \beta_{\text{op}}$
    \EndFor
    \State $\ell_{j_{\text{res}}^i} \gets \beta_{\text{res}}$
    
    \State \textbf{Step 4: Prior Construction}
    \If{$M = \text{Temp}$}
        \State $\pi_i \gets \mathrm{Softmax}(\bm{\ell} / \tau)$
    \ElsIf{$M = \text{Gumbel}$}
        \State Sample $g_j \sim \mathrm{Gumbel}(0,1)$ for each $j \in V$
        \State $\pi_i(j) \gets \frac{\exp((\ell_j + g_j)/\tau)}{\sum_{k \in V} \exp((\ell_k + g_k)/\tau)}$
    \ElsIf{$M = \text{Mix}$}
        \State $p_{\text{uniform}}(j) \gets \mathbb{I}[j \in \mathcal{K}_i] / |\mathcal{K}_i|$
        \State $p_{\text{result}}(j) \gets \mathbb{I}[j = j_{\text{res}}^i]$
        \State $\pi_i \gets \lambda \cdot p_{\text{uniform}} + (1-\lambda) \cdot p_{\text{result}}$
    \EndIf
    
    \State \textbf{Step 5: Selective Focus (Optional)}
    \If{using focused KL-divergence}
        \State Identify top-$k$ tokens $\mathcal{T}_i$ from $\pi_i$ with probability $> \delta$
        \State Retain only $\mathcal{T}_i$ for subsequent alignment
    \EndIf
\EndFor

\State \Return $\{\pi_i\}_{i=1}^N$

\end{algorithmic}
\end{algorithm*}

This appendix provides a detailed description of the Rule-Based Priors Construction pipeline, which transforms explicit reasoning chains into differentiable prior distributions for training our latent reasoning model. The complete procedure is formalized in Algorithm~\ref{alg:rule_based_priors}.

\subsection{Overview}

The pipeline comprises five main steps: (1) segmentation of explicit reasoning chains into computational steps, (2) extraction of operational and result tokens from each step, (3) initialization of logit vectors based on token categories, (4) construction of differentiable prior distributions using one of three methods, and (5) optional selective focus for KL-divergence computation. Each step is designed to preserve essential arithmetic semantics while ensuring differentiability for gradient-based optimization.

\subsection{Step-by-Step Description}

\textbf{Step 2: Token Extraction.} 
For each extracted step $r_i$, we identify two fundamental token categories:
\begin{itemize}
    \item \textbf{Operational tokens} $\mathcal{K}_i$: numerical values, arithmetic operators, and mathematical symbols that constitute the computational expression.
    \item \textbf{Result token} $j_{\text{res}}^i$: the numerical output produced by evaluating the step's expression.
\end{itemize}
This adaptive tokenization scheme preserves the essential arithmetic logic while accommodating natural variations in problem complexity.

\textbf{Step 3: Logit Initialization.} 
We initialize a logit vector $\bm{\ell} \in \mathbb{R}^{|V|}$ over the vocabulary $V$, setting $\ell_j = -\infty$ for all tokens $j$ not appearing in the step. For operational tokens in $\mathcal{K}_i$, we assign $\ell_j = \beta_{\text{op}}$, and for the result token $j_{\text{res}}^i$, we assign a higher value $\ell_j = \beta_{\text{res}}$ with $\beta_{\text{res}} > \beta_{\text{op}}$.

\textbf{Step 4: Prior Construction.} 
Based on the chosen method $M$, we construct differentiable prior distributions:
\begin{itemize}
    \item \textbf{Temperature-based prior} ($M = \text{Temp}$): $\pi_i = \mathrm{Softmax}(\bm{\ell} / \tau)$, where $\tau$ controls distribution entropy.
    \item \textbf{Gumbel-Softmax prior} ($M = \text{Gumbel}$): $\pi_i(j) = \frac{\exp((\ell_j + g_j)/\tau)}{\sum_{k \in V} \exp((\ell_k + g_k)/\tau)}$ with $g_j \sim \mathrm{Gumbel}(0,1)$, enabling differentiable categorical sampling.
    \item \textbf{Mixture prior} ($M = \text{Mix}$): $\pi_i = \lambda \cdot p_{\text{uniform}} + (1-\lambda) \cdot p_{\text{result}}$, where $p_{\text{uniform}}$ distributes probability uniformly over $\mathcal{K}_i$ and $p_{\text{result}}$ concentrates on $j_{\text{res}}^i$.
\end{itemize}

\textbf{Step 5: Selective Focus (Optional).} 
When using focused KL-divergence for training, we identify the top-$k$ tokens $\mathcal{T}_i$ from $\pi_i$ with probability above threshold $\delta$, retaining only these tokens for subsequent alignment. This reduces computational complexity from $O(|V|)$ to $O(k)$.

\subsection{Output}

The pipeline produces a sequence of prior distributions $\{\pi_i\}_{i=1}^N$, each corresponding to a reasoning step. These distributions serve as supervision signals for training the latent reasoner, enabling it to learn compressed distributional representations that preserve the symbolic structure and logical flow of the original reasoning process while maintaining differentiability for gradient-based optimization.

\end{document}